\documentclass[twocolumn]{svjour3}          
\smartqed  
\usepackage{graphicx}
\usepackage[numbers]{natbib}
\usepackage{times}
\usepackage{epsfig}
\usepackage{amsmath}
\usepackage{amssymb}
\usepackage{lineno}

\usepackage[pagebackref=true,breaklinks=true,letterpaper=true,colorlinks,bookmarks=false,citecolor=blue,linkcolor=blue]{hyperref}

\usepackage{booktabs}
\usepackage{amsmath}
\usepackage{amssymb}
\usepackage{verbatim}
\usepackage{multirow, makecell}
\usepackage{lineno}
\usepackage{enumerate}
\usepackage{amsfonts}
\usepackage{gensymb}
\usepackage{bm}
\usepackage{bbding}
\usepackage{comment}
\usepackage{color}
\usepackage{diagbox}
\usepackage{tabularx}
\usepackage{rotating}
\newcolumntype{L}[1]{>{\raggedright\let\newline\\\arraybackslash\hspace{0pt}}m{#1}}
\newcolumntype{C}[1]{>{\centering\arraybackslash}m{#1}}
\newcolumntype{R}[1]{>{\raggedleft\let\newline\\\arraybackslash\hspace{0pt}}m{#1}}
\newlength\savewidth\newcommand\shline{\noalign{\global\savewidth\arrayrulewidth
  \global\arrayrulewidth 1pt}\hline\noalign{\global\arrayrulewidth\savewidth}}
\newcommand{\tablestyle}[2]{\setlength{\tabcolsep}{#1}\renewcommand{\arraystretch}{#2}\centering\footnotesize}

\usepackage{colortbl}
\definecolor{Gray}{gray}{0.9}
\definecolor{GrayT}{gray}{0.4}
\newcommand{\etal}{{\em et al.}}
\newcommand{\eg}{{\em e.g.}}
\newcommand{\ie}{{\em i.e.}}

\usepackage{subcaption}
\usepackage[labelfont=bf]{caption}
\definecolor{light-blue}{RGB}{186,175,255}
\definecolor{light-red}{RGB}{255,137,165}
\usepackage{arydshln} 
\usepackage{booktabs}
\usepackage{multirow}
\usepackage{amsmath}
\usepackage{etoolbox}
\usepackage{bm}
\usepackage{mathtools}
\usepackage{algorithm}
\usepackage{algpseudocode}
\usepackage{xspace}

\usepackage{multirow, makecell}
\usepackage{colortbl}  
\usepackage{xcolor}
\usepackage{array}
\definecolor{Gray}{gray}{0.94}
\definecolor{liGray}{gray}{0.5}
\definecolor{LightCyan}{rgb}{0.88,1,1}
\usepackage{mathrsfs}

\begin{document}
\begin{sloppypar}

\title{CLIP-guided Prototype Modulating for Few-shot Action Recognition
}


\author{Xiang Wang      \and
        Shiwei Zhang    \and
        Jun Cen       \and
        Changxin Gao       \and
        Yingya Zhang \and \\
     Deli Zhao \and
     Nong Sang
}


\institute{
Xiang Wang  \and Changxin Gao  \and Nong Sang (Corresponding author)
\at
              Key Laboratory of Ministry of Education for Image Processing and Intelligent Control, School of Artificial
Intelligence and Automation, Huazhong University of Science and Technology  \\
              \email {\{wxiang, cgao, nsang\}@hust.edu.cn}           
           \and
           Shiwei Zhang \and Yingya Zhang \and  Deli Zhao \at
              Alibaba Group \\
              \email {zhangjin.zsw@alibaba-inc.com, yingya.zyy@alibaba-inc.com  and zhaodeli@gmail.com}
          \and
          Jun Cen \at 
          The Hong Kong University of Science and Technology \\
          \email {jcenaa@connect.ust.hk}
}

\date{Received: date / Accepted: date}

\maketitle

\begin{abstract}
Learning from large-scale contrastive language-image pre-training like CLIP has shown remarkable success in a wide range of downstream tasks recently, but it is still under-explored on the challenging few-shot action recognition (FSAR) task.
In this work, we aim to transfer the powerful multimodal knowledge of CLIP to alleviate the inaccurate prototype estimation issue due to data scarcity, which is a critical problem in low-shot regimes.
%
To this end, we present a CLIP-guided prototype modulating framework called CLIP-FSAR, which consists of two key components: a video-text contrastive objective and a prototype modulation. 
%
%
Specifically, 
the former bridges the task discrepancy between CLIP and the few-shot video task by contrasting videos and corresponding class text descriptions.
The latter leverages the transferable textual concepts from CLIP to adaptively refine visual prototypes with a temporal Transformer.
%
By this means, CLIP-FSAR can take full advantage of the rich semantic priors in CLIP to obtain reliable prototypes and achieve accurate few-shot classification.
%
Extensive experiments on five commonly used benchmarks demonstrate the effectiveness of our proposed method, and CLIP-FSAR significantly outperforms existing state-of-the-art methods under various settings.
The source code and models will be publicly available at \url{https://github.com/alibaba-mmai-research/CLIP-FSAR}.

\keywords{Few-shot Action Recognition  \and Multimodal Learning  \and Multimodal Foundation Models \and Large Model Application}

\end{abstract}

\section{Introduction}
\label{sec:intro}

Action recognition is a fundamental topic in video understanding field and has made significant progress in recent years~\cite{TSM,vivit,Kinetics,wang2021self,wang2021oadtr}.
Despite this, modern models require massive data annotation, which may be time-consuming and laborious to collect.
Few-shot action recognition is a promising direction to alleviate the data labeling problem, which aims to identify unseen classes with a few labeled videos and has received considerable attention~\cite{CMN,OTAM,HyRSM}.

Mainstream few-shot action recognition approaches follow the metric-based meta-learning paradigm~\cite{MatchNet} to optimize the model, which first maps the input videos into a common feature space and then calculates the alignment distances from the query to the support prototypes through 
pre-defined metric rules for classification.
To obtain a discriminative feature space and facilitate the learning, existing methods generally take advantage of a single-modal supervised pre-training, \eg, ImageNet initialization~\cite{imagenet}.
\begin{figure}[t]
  \centering
   \includegraphics[width=0.99\linewidth]{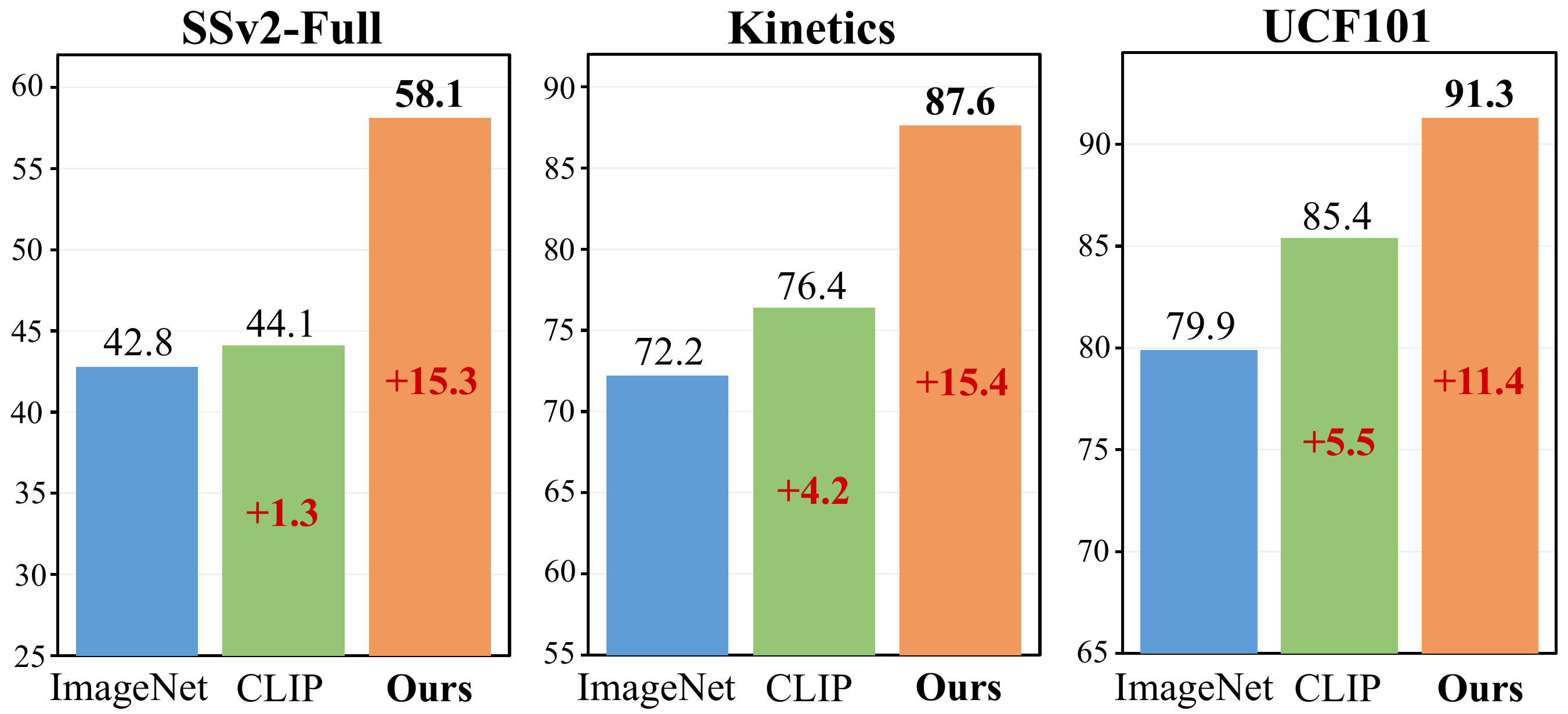}
\vspace{-1mm}
   \caption{
   Performance comparison based on a typical few-shot action recognition method, \ie, OTAM~\cite{OTAM}.
   We notice that directly replacing the original ImageNet pre-trained backbone with the visual encoder of CLIP~\cite{CLIP} and simply finetuning the CLIP model can only yield limited performance improvements.
   In contrast, our approach provides significant performance gains by fully exploiting CLIP's multimodal semantic knowledge.
   }
   \label{fig:Motivation}
   \vspace{-0mm}
\end{figure}
%
%
Despite the remarkable progress, the single-modal initialization still cannot achieve a satisfactory performance due to relatively limited labeled data and a lack of multimodal correspondence information.
Fortunately, contrastive language-image pre-training is an powerful emerging paradigm to learn high-quality transferable representations by 
applying
contrastive learning from large-scale image-text pairs available on the Internet~\cite{CLIP,ALIGN}.
Typical work like CLIP~\cite{CLIP} successfully demonstrates amazing transferability and has made remarkable progress on various downstream tasks~\cite{coop,Clip-adapter,Tip-adapter,Denseclip,wang2022cris}.
Inspired by these successful practices, we attempt to transfer the powerful capability of CLIP to address the few-shot action recognition task.
A straightforward idea is to replace the ImageNet initialization with the pre-trained visual encoder of CLIP to exploit the strong representation capacity.
However, experimental results in Figure~\ref{fig:Motivation} indicate that naive replacement of the initialization and finetuning on downstream datasets can indeed improve performance but is still limited.
We attribute this to insufficiently exploiting the multi-modal properties of the CLIP model to obtain reliable support prototypes in few-shot
scenarios.
%

%
%
Motivated by the above observations, we focus on fully leveraging the powerful multimodal knowledge of the CLIP model for few-shot action recognition and thus propose a novel CLIP-guided prototype modulating framework, termed CLIP-FSAR.
%
%
Specifically, to adapt CLIP to the few-shot video task, we optimize a video-text contrastive objective to pull close the video features and the corresponding class text representations.
%
%
%
Moreover, to alleviate the dilemma of insufficient visual information in the few-shot scenario, we propose to modulate the visual prototypes with the guidance of textual semantic priors in CLIP.
This is achieved by implementing a temporal Transformer to adaptively fuse
the textual and visual features in the support set.
%
%
%
%
Based on this scheme, our CLIP-FSAR enables the generation of comprehensive and reliable prototypes for the few-shot metric objective, yielding robust classification results.
We conduct extensive experiments on five standard benchmarks, and the results demonstrate that our CLIP-FSAR significantly improves the baseline methods.
In summary, our major contributions are as follows:
(1) We propose a novel CLIP-FSAR for few-shot action recognition, which takes full use of the multimodal knowledge of the CLIP model. 
To the best of our knowledge, this is the first attempt to apply the large-scale contrastive language-image pre-training to the few-shot action recognition field.
%
(2) We design a video-text contrastive objective for CLIP adaptation and a prototype modulation to generate reliable prototypes. 
(3) Extensive experiments on five challenging benchmarks demonstrate the effectiveness of our approach, and  CLIP-FSAR achieves state-of-the-art performance.

\section{Related work}
\label{sec:related}

This work is closely relevant to few-shot image classification, vision-language contrastive learning, and few-shot action recognition. 
We give a brief overview of them below.

\vspace{1mm}
\noindent \textbf{Few-shot image classification.} 
The goal of few-shot learning is to identify new classes with a small number of samples~\cite{few-shot_feifei,guo2020broader}. 
The few-shot research can be broadly divided into three groups: data augmentation, model optimization, and metric-based methods.
The first class of methods~\cite{data-aug-3,data-aug-1,data-aug-6,ye2021learning} attempts to use data-generating strategies to improve the generalization of few-shot models.
The second hopes to provide a good initialization for the classification model so that only a small number of gradient update steps are required to reach the optimum point.
Typical methods include MAML~\cite{MAML} and other variants~\cite{MAML-1,MAML-2,MAML-3,MAML-4,MAML-5}.
Recently, ~\cite{coop,Clip-adapter} also try to fine-tune part of the learnable weights at test time to transfer the robust prior of CLIP~\cite{CLIP} for few-shot classification.
Metric-based methods classify query samples by learning a mapping space and setting support-query matching rules, including cosine similarity~\cite{MatchNet,ye2020few}, Euclidean distance~\cite{prototypical,yoon2019tapnet}, and learnable metrics~\cite{RelationNet,Finding_task-relevant}.
There are also some methods~\cite{pahde2019self,pahde2021multimodal,xing2019adaptive,xu2022attribute} that introduce additional attribute information to assist with few-shot classification.
Among these, our approach falls under the
metric-based line that classifies query samples without any test time finetuning 
and tries to leverage the transferable multimodal knowledge of CLIP to boost the challenging few-shot action recognition task.

\vspace{1mm}
\noindent \textbf{Vision-language contrastive learning.} 
Exploring visual-language pre-training is a remarkably hot and promising direction to exploit the association of different attribute data~\cite{li2020oscar,li2021align,li2020unimo,wang2022vlmixer,wang2021actionclip,zhai2022lit,zhong2022regionclip,yang2022unified}.
Recently, contrastive language-image  pre-training~\cite{CLIP,ALIGN,luo2021coco} has received increasing attention due to its simplicity and effectiveness.
Representative works such as CLIP~\cite{CLIP} project images and natural language descriptions to a common feature space through two separate encoders for contrastive learning, and achieve significant ``zero-shot" transferability by pre-training on hundreds of millions of image-text pairs.
%
Subsequently, these pre-trained models have been extended to various downstream tasks and shown excellent performance, including image classification~\cite{coop,CoCOOP}, object detection~\cite{Proposalclip,gu2021open}, semantic segmentation~\cite{wang2022cris,Denseclip}, and video understanding~\cite{lin2022frozen,ju2021prompting,liu2022ts2}.
%
%
%
%
Inspired by these successes, 
in this work we present the first simple but efficient framework to leverage the rich semantic knowledge of CLIP for few-shot action recognition.

\vspace{1mm}

\noindent \textbf{Few-shot action recognition.} 
The majority of existing few-shot recognition approaches follow the metric-based manner~\cite{MatchNet} to optimize the model and design robust alignment metrics to calculate
the distances between the query and support samples for classification.
Some methods~\cite{zhu2021few,ARN-ECCV,CMN,CMN-J} adopt the idea of global matching in the field of few-shot image classification~\cite{prototypical,RelationNet} to carry out few-shot matching, which results in relatively poor performance because long-term temporal alignment information is ignored in the measurement process.
To exploit the temporal cues,
the following approaches~\cite{OTAM,ITANet,TRX,TA2N,MTFAN,STRM,HyRSM,nguyen2022inductive,huang2022compound,wang2023hyrsm++,HCL} focuses on local frame-level (or segment-level)
alignment between query and support videos.
Among them, OTAM~\cite{OTAM} proposes a variant of the dynamic time warping technique~\cite{DTW} to explicitly utilize
the temporal ordering information in support-query video pairs. 
ITANet~\cite{ITANet} designs a hybrid spatial and channel attention mechanism to learn representative features.
TRX~\cite{TRX} exhausts ordered tuples of varying numbers of frames to compare support and query videos.
HyRSM~\cite{HyRSM} introduces self-attention to aggregate temporal relations among videos and designs a bidirectional Mean Hausdorff Metric (Bi-MHM) to relax temporal constraints in the matching process.
Huang \etal~\cite{huang2022compound} propose compound prototypes, which adopt multiple metric functions for few-shot matching.
\begin{figure*}[t]
  \centering
   \includegraphics[width=0.94\linewidth]{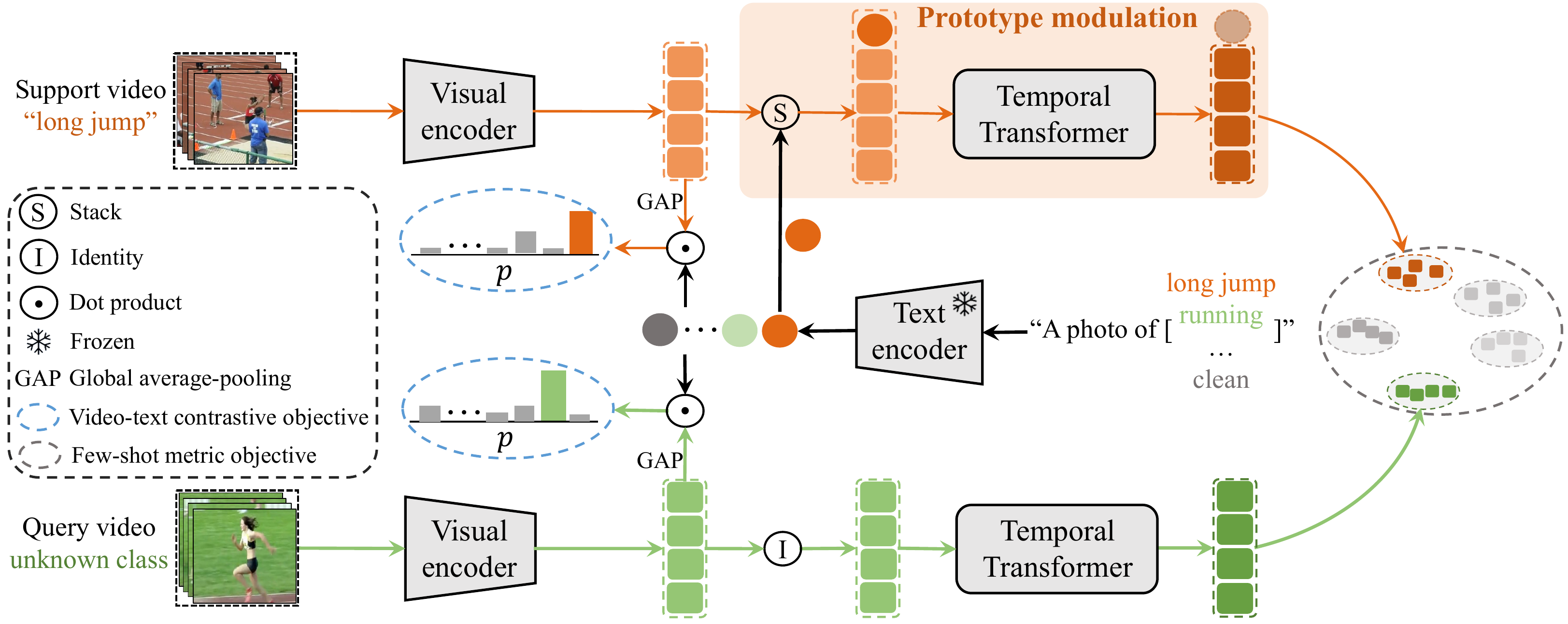}
\vspace{-3mm}
   \caption{
   The architecture of CLIP-FSAR.
   Given the support and query videos, a visual encoder is first adopted to extract the 
   frame features.
   Then, we apply a video-text contrastive objective to pull close the obtained video features and the corresponding class text representations.
   Afterward, a prototype modulation is employed to refine the visual support prototypes with textual semantic features.
   Finally, a few-shot metric objective is imposed on the resulting informative prototypes and query features for classification.
   }
   \label{fig:network}
   \vspace{-2mm}
\end{figure*}
%
%
Even with great success, these methods mainly employ single-modal pre-trained backbones without involving multimodal features, which limits further performance gains. 
In our work, we introduce the powerful  CLIP model and attempt to exploit its multimodal semantic knowledge to handle the few-shot action recognition task.
Note that there are two essential differences between the proposed CLIP-FSAR and the original linear-probe CLIP~\cite{CLIP}:
1) The original linear-probe CLIP~\cite{CLIP} performs a linear-probe evaluation for action recognition and needs to finetune on the test classes, which is completely different from our metric-based few-shot setup, where the training and testing classes are disjoint and finetuning on test classes is not allowed;
2) The major contribution is to design an effective method that can leverage CLIP's multimodal knowledge to generate reliable prototypes, while the original linear-probe CLIP just linear-finetunes the visual backbone without involving multimodal cues.
In addition, in this paper our key insight is that the naive extension of CLIP to the few-shot action recognition task is obviously inadequate (as shown in Figure~\ref{fig:Motivation}), thus we carefully design CLIP-FSAR to better leverage the powerful multimodal knowledge of CLIP and hope to provide the community with a new and feasible solution that leverages large foundation models and fine-tunes on downstream task.
%

\section{Methodology}
\label{sec:method}

We first briefly review the background knowledge about CLIP~\cite{CLIP} and then describe in detail how our framework applies CLIP to the few-shot action recognition task.

\subsection{Preliminaries of CLIP}

%
%
Unlike the traditional paradigm of supervised training using human-annotated labels~\cite{Resnet,imagenet,Kinetics}, Contrastive Language-Image Pre-training (CLIP)~\cite{CLIP} employs natural language descriptions to supervise representation learning, which improves the scalability and transferability of the learned model.
More specifically, 
CLIP is a dual-encoder structure consisting of a visual encoder and a text encoder.
%
The visual encoder is intended to map the input images into a compact embedding space and can be implemented with the ResNet~\cite{Resnet} architecture or ViT~\cite{ViT} architecture.
The text encoder is designed to extract high semantic features from natural language descriptions based on Transformers~\cite{Transformer}.
Subsequently, the obtained image features and text features are pulled closer if they are matched pairs and pushed away if not. 
In a word, the learning objective of CLIP is to perform multimodal contrastive learning in the common feature space.
To further improve the representation ability, CLIP is pre-trained on 400 million web-crawled image-text pairs.
When transferring to downstream tasks (\eg, object detection~\cite{gu2021open}), in order to align with the textual description during pre-training, the input text generally utilizes the prompt template ``\texttt{a photo of [CLS]}'', where \texttt{[CLS]} denotes the actual category name.

\subsection{CLIP-FSAR}

\textbf{Overview.}
Few-shot action recognition aims to recognize new action classes with a small number of videos.
In a typical few-shot setting, there are two class disjoint datasets, a base dataset $\mathcal{D}_{train}$ for training and a novel dataset $\mathcal{D}_{test}$ for testing.
In order to simulate the test environment, a large number of few-shot tasks (or called episodes) are generally sampled from $\mathcal{D}_{train}$ during the training process to optimize the model.
For a standard $N$-way $K$-shot task,  there is a support set $S=\{s_{1},s_{2},...,s_{N \times K} \}$ consisting of $N$ classes and $K$ videos per class.
The objective of the task is to classify the query video $q$ based on the support samples.
For the convenience of formulation, we consider the $N$-way 1-shot (\ie, $K=1$) task to present our framework.
Following previous works~\cite{OTAM,TRX,ITANet}, a sparse frame sampling strategy~\cite{TSN} is implemented on the input video to reduce the computational burden.
We first adopt the visual encoder $\mathcal{V}$  of CLIP to generate the features of input video frames and the text encoder $\mathcal{T}$ to extract the textual embeddings of the corresponding class natural language descriptions.
Then we apply a video-text contrastive objective over these obtained frame features and text features to adapt CLIP to the few-shot video task.
Moreover, we propose a prototype modulation to refine the visual prototypes for the few-shot metric objective.
Figure~\ref{fig:network} exhibits the entire framework of our CLIP-FSAR.

\textbf{Video-text contrastive objective.}
Given a support set $S=\{s_{1},s_{2},...,s_{N} \}$ and a query video $q=\{q^{1},q^{2},...,q^{t}\}$, where $s_{i}=\{s^{1}_{i},s^{2}_{i},...,s^{t}_{i} \}$ is a support video consisting of sparsely sampled $t$ frames.
The visual encoder $\mathcal{V}$ is employed to encode the video features:
\begin{equation}
\centering
    f_{s_{i}} = \mathcal{V}(s_{i}), \quad f_{q} = \mathcal{V}(q)
\end{equation}
where $f_{s_{i}} \in \mathbb{R}^{t\times C}$,  $f_{q} \in \mathbb{R}^{t\times C}$, and $C$ is the number of channels.
We then use the text encoder $\mathcal{T}$ to extract the textual features of base classes with the prompt template ``\texttt{a photo of [CLS]}'' and denote the obtained textual features as $\{w_{i}\}_{i=1}^{B}$, where $B$ is the total class number of base set $\mathcal{D}_{train}$ and $w_{i} \in \mathbb{R}^{C}$ is a feature vector.
Following previous practices~\cite{wang2022cris,Proposalclip,Denseclip},
in order to preserve the original pre-trained transferable knowledge in the text encoder and reduce the optimization burden, we fix $\mathcal{T}$ not to update during training.
To bridge the task discrepancy between CLIP and the few-shot video task, we simulate the original CLIP training objective to maximize the similarity of video features and text features if they are matched pairs and minimize otherwise.
To achieve this goal, we first compute the video-text matching probability as follows:
\begin{equation}
\label{video-text-eq}
    p^{video-text}_{(y=i| v)}  = \frac{\exp (\mathrm{sim}(\mathrm{GAP}(f_{v}), w_{i})/\tau )}{{\textstyle \sum_{j=1}^{B} \exp (\mathrm{sim} ((\mathrm{GAP}(f_{v}),w_{j} )/\tau )} } 
\end{equation}
where $v\in \{s_{1},s_{2},...,s_{N},q\}$, $\mathrm{sim}$ is a cosine similarity function, $\mathrm{GAP}$ is short for global average-pooling,   and $\tau$ denotes a learnable temperature factor.
Then we impose a cross-entropy loss $\mathcal{L}_{video-text}$ between the predictions and the actual class label to optimize the objective.

\textbf{Prototype modulation.}
Existing few-shot action recognition methods, such as OTAM~\cite{OTAM}, generally classify query video $q$ by comparing its temporal alignment distance from support visual prototypes.
In the few-shot action recognition task, video prototype is a sequence of frame prototypes, please refer to OTAM~\cite{OTAM} for more details.
The distance between the query video $q$ and the support video $s_{i}$ can be  expressed as:
\begin{equation}
    d_{q,s_{i}} = \mathcal{M}(f_{q}, f_{s_{i}})
\end{equation}
where $\mathcal{M}$ stands for a temporal alignment metric.
In OTAM, $\mathcal{M}$ is a variant of the dynamic time warping~\cite{DTW} to measure the support-query distance as the frame alignment cost.
%
%
%
The few-shot performance heavily relies on the accuracy of prototype estimation~\cite{prototypical}.
%
On the one hand, in low-shot scenarios,  visual information tends to be insufficient due to data scarcity, leading to inaccurate prototypes.
On the other hand, there is complementarity between visual and textual modalities~\cite{shi2021multi}, and the textual description representations of CLIP involve rich semantic priors~\cite{li2021language}.
%
%
Based on these,
to boost the reliability of the support prototypes, we propose to utilize the informative support text features to refine the prototypes.
Concretely, on the basis of the support visual feature $f_{s_{i}}$, we stack the text features to the corresponding video $s_{i}$ along the temporal dimension, \ie, $\mathbb{R}^{t\times C} \cup  \mathbb{R}^C \rightarrow \mathbb{R}^{(t+1)\times C}$, and use a temporal Transformer to adaptively fuse the features.
We denote the resulting enhanced visual features as $\widetilde{f}_{s_{i}}$ (excluding the output text feature). 
Since we do not know the real class information of the query video during testing, we only input the visual query feature into the temporal Transformer so that the output query feature $\widetilde{f}_{q}$ and support features can be matched in a common feature space.
Subsequently, we adopt the temporal alignment metric to calculate the query-support distance:
\begin{equation}
    d^{\prime }_{q,s_{i}} = \mathcal{M}(\widetilde{f}_{q}, \widetilde{f}_{s_{i}})
\end{equation}
where $\mathcal{M}$ is the OTAM metric by default in our CLIP-FSAR.
Note that the proposed CLIP-FSAR is a plug-and-play framework and
in the later experimental section, we will insert CLIP-FSAR into other existing metrics or methods~\cite{ITANet,TRX,HyRSM} and empirically demonstrate its pluggability.
Based on the distances,
the probability distribution
over support classes for the query video $q$ can be formulated as:
\begin{equation}
\label{few-shot-eq}
    p^{few-shot}_{(y = i | q) } = \frac{\exp ( d^{\prime }_{q,s_{i}})}{{\textstyle \sum_{j=1}^{N} \exp (d^{\prime }_{q,s_{j}} )} } 
\end{equation}
Following previous works~\cite{OTAM,HyRSM,TRX}, we can use a cross-entropy loss $\mathcal{L}_{few-shot}$ to optimize the model parameters.
The final training objective of our CLIP-FSAR is:
\begin{equation}
\label{vt-eq}
   \mathcal{L} = \mathcal{L}_{video-text} + \alpha \mathcal{L}_{few-shot}
\end{equation}
where $\alpha$ is a balance factor.
For few-shot evaluation, we can obtain the matching probabilities belonging to the support classes by Equation~\ref{few-shot-eq}, like previous works~\cite{OTAM,TRX,HyRSM}.
In addition, due to the two-objective design of our proposed framework, we can also combine the video-text matching result (Equation~\ref{video-text-eq}) and the few-shot classification result (Equation~\ref{few-shot-eq}) to obtain the merged prediction:
\begin{equation}
\label{combine-eq}
    p^{\dagger}_{(y=i | q )} = ( p^{video-text}_{(y=i| q)})^{\beta } \cdot (p^{few-shot}_{(y = i | q) })^{1-\beta }
\end{equation}
where $\beta \in [0, 1]$ is an adjustable hyperparameter, and we denote the above ensemble manner as CLIP-FSAR$^{\dagger}$.
Note that the combination of zero-shot and few-shot results described above is only an optional approach, and we primarily focus on the few-shot performance in this paper.

\begin{table*}[t]
\centering
\footnotesize
\tablestyle{5pt}{1.1}
\caption{Comparison to recent few-shot action recognition approaches on SSv2-Full and Kinetics. 
The experiments are conducted under the 5-way  $K$-shot setting with $K$ changing from 1 to 5.
``-" means the result is not available in published works.
``${*}$"  indicates the results of our implementation. 
``INet-RN50" denotes ResNet-50 pretrained on ImageNet.
%
``${\dagger}$" represents the ensemble method in Equation~\ref{combine-eq}, which combines the few-shot results and video-text matching results.
}
\vspace{-3mm}
\setlength{\tabcolsep}{1.6mm}{
\begin{tabular}
{l|c|c|ccccc|ccccc}
			
\hspace{-0.5mm}  \multirow{2}{*}{Method} \hspace{2mm} & \multicolumn{1}{c|}{\multirow{2}{*}{Reference}} &\multicolumn{1}{c|}{\multirow{2}{*}{Pre-training}} & \multicolumn{5}{c|}{{SSv2-Full}}  & \multicolumn{5}{c}{{Kinetics}}  \\
& \multicolumn{1}{c|}{} & \multicolumn{1}{c|}{} 
& \multicolumn{1}{l}{1-shot} & 2-shot  & 3-shot  & 4-shot   & 5-shot 
&  \multicolumn{1}{l}{1-shot} & 2-shot  & 3-shot  & 4-shot   & 5-shot  \\ \shline

MatchingNet~\cite{MatchNet}    & { NeurIPS'16} & INet-RN50
& -  & -  & -   & -   & -   
& 53.3   & 64.3  & 69.2  & 71.8  & 74.6  \\
MAML~\cite{MAML}    & ICML'17& INet-RN50
& -  & -  & -   & -   & -   
& 54.2   & 65.5  & 70.0  & 72.1  & 75.3  \\
Plain CMN~\cite{CMN}    & ECCV'18& INet-RN50
& -  & -  & -   & -   & -   
& 57.3   & 67.5  & 72.5  & 74.7  & 76.0  \\
CMN++~\cite{CMN}    & ECCV'18& INet-RN50
& 34.4  & -  & -   & -   & 43.8   
& -   & -  & -  & -  & -  \\
TRN++~\cite{TRN-ECCV}    & ECCV'18& INet-RN50
& 38.6  & -  & -   & -   & 48.9   
& -   & -  & -  & -  & -  \\
TARN~\cite{TARN}    & BMVC'19& INet-RN50
& -  & -  & -   & -   & -   
& 64.8   & -  & -  & -  & 78.5  \\
CMN-J~\cite{CMN-J}    & TPAMI'20& INet-RN50
& -  & -  & -   & -   & -   
& 60.5   & 70.0  & 75.6  & 77.3  & 78.9  \\
ARN~\cite{ARN-ECCV}    & ECCV'20& -
& -  & -  & -   & -   & -   
& 63.7   & -  & -  & -  & 82.4  \\
OTAM~\cite{OTAM}    & CVPR'20& INet-RN50
& 42.8  & 49.1  & 51.5   & 52.0   & 52.3   
& \hspace{0.6mm} 72.2$^{*}$   & 75.9  & 78.7  & 81.9  &  \hspace{0.6mm} 84.2$^{*}$  \\
ITANet~\cite{ITANet}    & IJCAI'21& INet-RN50
& 49.2  & 55.5  & 59.1   & 61.0   & 62.3   
& 73.6   & -  & -  & -  & 84.3  \\
%
TRX  ($\{1\}$)   & CVPR'21& INet-RN50     & 38.8     & 49.7   & 54.4     & 58.0      & 60.6   & 63.6  & 75.4  & 80.1  & 82.4  & 85.2   \\ 
TRX ($\{2,3\}$)~\cite{TRX}  & CVPR'21  & INet-RN50        &  42.0   &  53.1   &  57.6   &   {61.1}   &   {64.6}   & 63.6  & {76.2}  & {81.8}  & {83.4}  & {85.9}  \\  
TA$^{2}$N~\cite{TA2N}    & AAAI'22& INet-RN50
& 47.6  & -  & -   & -   & 61.0   
& 72.8   & -  & -  & -  & 85.8  \\
MTFAN~\cite{MTFAN}    & CVPR'22& INet-RN50
& 45.7  & -  & -   & -   & 60.4   
& {74.6}   & -  & -  & -  & {87.4}  \\
STRM~\cite{STRM}    & CVPR'22& INet-RN50
& 43.1 & 53.3 & 59.1 & 61.7 & 68.1
& 62.9 & 76.4 & 81.1 & 83.8 & 86.7 \\
HyRSM~\cite{HyRSM}    & CVPR'22& INet-RN50
& 54.3  & {62.2}  & {65.1}   & {67.9}   & 69.0   
& 73.7   & 80.0  & {83.5}  & {84.6}  & 86.1  \\
Nguyen~\etal~\cite{nguyen2022inductive}    & ECCV'22& INet-RN50
& 43.8  & -  & -   & -   & 61.1   
& {74.3}   & -  & -  & -  & {87.4}  \\

Huang~\etal~\cite{huang2022compound}    & ECCV'22 & INet-RN50
& 49.3  & -  & -   & -   & 66.7   
& 73.3   & -  & -  & -  & 86.4 \\
HCL~\cite{HCL}    & ECCV'22 & INet-RN50
& 47.3  & 54.5  & 59.0   & 62.4   & 64.9   
& 73.7   & 79.1  & 82.4  & 84.0  & 85.8 
\\ \shline
OTAM (Baseline)~\cite{OTAM}    & CVPR'20 & CLIP-RN50
& 44.1  & 52.4  & 55.6   & 58.1   & 59.8   
& 76.4   & 84.0  & 86.1  & 87.7  & 88.9  \\
CLIP-Freeze~\cite{CLIP}    & ICML'21 & CLIP-RN50
& 28.5  & 32.5  & 34.5   & 37.1   & 37.8   
& 68.2   & 78.3  & 82.1  & 84.0  & 85.3  \\
\rowcolor{Gray}
\textbf{CLIP-FSAR}  & -& CLIP-RN50
& {58.1}  & 59.0  & 60.7   & 61.4   & 62.8  
& 87.6   & 89.3  & 90.7  & 91.7  & 91.9  \\ 
\rowcolor{Gray}
\rowcolor{Gray}
\textcolor{liGray}{\textbf{CLIP-FSAR$^{\dagger}$}} & \textcolor{liGray}{-}& \textcolor{liGray}{CLIP-RN50}
& \textcolor{liGray}{58.7}  & \textcolor{liGray}{59.2}  & \textcolor{liGray}{60.7}   & \textcolor{liGray}{61.4}   & \textcolor{liGray}{62.8}  
& \textcolor{liGray}{90.1}   & \textcolor{liGray}{90.4}  & \textcolor{liGray}{90.8}  & \textcolor{liGray}{91.9}  & \textcolor{liGray}{92.0}  \\ 
\shline
OTAM (Baseline)~\cite{OTAM}    & CVPR'20 & CLIP-ViT-B
& 50.2  & 57.4  & 63.8   & 65.5   & 68.6   
&  88.2 & 92.4  & 93.6   & 94.2   & 94.8    \\
CLIP-Freeze~\cite{CLIP}    & ICML'21 & CLIP-ViT-B
&  30.0 & 34.4  & 38.2   & 40.6   & 42.4   
& 78.9  & 87.0  & 90.2   & 91.4   & 91.9    \\
\rowcolor{Gray}
\textbf{CLIP-FSAR} & -& CLIP-ViT-B
& \textbf{61.9}  & \textbf{64.9}  & \textbf{68.1}   & \textbf{70.9}   & \textbf{72.1}    
&  \textbf{89.7} & \textbf{92.9}  & \textbf{94.2}   & \textbf{94.8}   & \textbf{95.0}    \\ 
\rowcolor{Gray}
\rowcolor{Gray}
\textcolor{liGray}{\textbf{CLIP-FSAR$^{\dagger}$}}  & \textcolor{liGray}{-}& \textcolor{liGray}{CLIP-ViT-B}
&  \textcolor{liGray}{62.1} & \textcolor{liGray}{65.0}  & \textcolor{liGray}{68.3}   & \textcolor{liGray}{70.9}   & \textcolor{liGray}{72.1}    
& \textcolor{liGray}{94.8}  & \textcolor{liGray}{95.0}  & \textcolor{liGray}{95.0}   & \textcolor{liGray}{95.1}   & \textcolor{liGray}{95.4}     \\ 
%
%
\end{tabular}
}

\label{tab:compare_SOTA_1}
\vspace{-3mm}
\end{table*}

\section{Experiments}
\label{sec:exper}

In this section, we conduct extensive experiments on five few-shot action recognition benchmarks and compare CLIP-FSAR with current state-of-the-art methods to verify its effectiveness.
Besides, detailed ablation studies are performed to analyze the properties of CLIP-FSAR.

\subsection{Experimental settings}

\noindent \textbf{Datasets.}
We validate our approach on five commonly used few-shot action datasets, including SSv2-Full~\cite{SSv2}, SSv2-Small~\cite{SSv2}, Kinetics~\cite{Kinetics}, UCF101~\cite{UCF101}, and HMDB51~\cite{HMDB51}.
Referring to the setups in ~\cite{OTAM,CMN,TRX,ITANet}, on SSv2-Full, SSv2-Small, and Kinetics, we take randomly sampled 64 classes as base classes and 24 classes as novel classes for a fair comparison.
For UCF101 and HMBD51, we follow the dataset splits from ~\cite{ARN-ECCV} to evaluate our CLIP-FSAR.
%


%
\begin{table*}[ht]
\centering
\small
\tablestyle{6pt}{1.2}
\caption{Comparison to recent few-shot action recognition methods on the UCF101, SSv2-Small  and HMDB51 datasets. 
The experiments are conducted under the 5-way $K$-shot setting with $K$ changing from 1 to 5.
"-" means the result is not available in published works.
``${*}$"  stands for the results of our implementation.
``INet-RN50" denotes ResNet-50 pretrained on ImageNet.
``${\dagger}$" represents the ensemble method in Equation~\ref{combine-eq}, which combines the few-shot results and video-text matching results.
}
\vspace{-3mm}
\begin{tabular}
{l|c|c|ccc|ccc|ccc}
			
\hspace{-0.5mm}  \multirow{2}{*}{Method} \hspace{2mm} & \multicolumn{1}{c|}{\multirow{2}{*}{Reference}} &
\multicolumn{1}{c|}{\multirow{2}{*}{Pre-training}} &\multicolumn{3}{c|}{{UCF101}}  &\multicolumn{3}{c|}{{SSv2-Small}}  & \multicolumn{3}{c}{{HMDB51}}  \\
& \multicolumn{1}{c|}{} & \multicolumn{1}{c|}{} 
& \multicolumn{1}{l}{1-shot} & 3-shot  & 5-shot & \multicolumn{1}{l}{1-shot} & 3-shot   & 5-shot 
& \multicolumn{1}{l}{1-shot}& 3-shot    & 5-shot  \\ \shline
MatchingNet~\cite{MatchNet}    & NeurIPS'16 & INet-RN50
& -  & -  & -   
& 31.3   & 39.8   & 45.5   
& -  & -  & -    \\
MAML~\cite{MAML}    & ICML'17& INet-RN50
& -  & -  & -   
& 30.9   & 38.6   & 41.9   
& -  & -  & -    \\
Plain CMN~\cite{CMN}    & ECCV'18& INet-RN50
& -  & -  & -   
& 33.4   & 42.5   & 46.5   
& -  & -  & -    \\
CMN-J~\cite{CMN-J}    & TPAMI'20& INet-RN50
& -  & -  & -   
& 36.2   & 44.6   & 48.8   
& -  & -  & -    \\
ARN~\cite{ARN-ECCV}    & ECCV'20& -
& 66.3  & -  & 83.1   
& -   & -   & -   
& 45.5  & -  & 60.6    \\
OTAM~\cite{OTAM}    & CVPR'20& INet-RN50
& 79.9  & 87.0  & 88.9   
& 36.4  & 45.9  & 48.0   
& 54.5  & 65.7  & 68.0    \\
ITANet~\cite{ITANet}    & IJCAI'21& INet-RN50
& -  & -  & -   
& 39.8  & 49.4  & 53.7  
& -  & -  & -    \\
TRX~\cite{TRX}    & CVPR'21& INet-RN50
& 78.2  & 92.4  & {96.1}   
& 36.0  & 51.9  & \hspace{0.6mm} 56.7$^{*}$  
& 53.1  & 66.8  & 75.6    \\
TA$^{2}$N~\cite{TA2N}    & AAAI'22& INet-RN50
& 81.9  & -  & 95.1   
& -  & -  & -   
& 59.7  & -  & 73.9    \\
MTFAN~\cite{MTFAN}    & CVPR'22& INet-RN50
& 84.8  & -  & 95.1   
& -  & -  & -  
& 59.0  & -  & 74.6    \\
STRM~\cite{STRM}    & CVPR'22& INet-RN50
& 80.5 & 92.7 & {96.9}
& 37.1 & 49.2 & 55.3
& 52.3 & 67.4 & {77.3}   \\
HyRSM~\cite{HyRSM}    &  CVPR'22& INet-RN50
& 83.9  & 93.0  & 94.7   
& 40.6  & {52.3}  & 56.1  
& {60.3}  & {71.7}  & 76.0    \\
Nguyen~\etal~\cite{nguyen2022inductive}    & ECCV'22& INet-RN50
& 84.9  & -  & 95.9   
& -  & -  & -  
& 59.6  & -  & {76.9}    \\
Huang~\etal~\cite{huang2022compound}    & ECCV'22& INet-RN50
& 71.4  & -  & 91.0   
& 38.9  & -  & {61.6}  
& 60.1  & -  & {77.0}    \\ 
HCL~\cite{HCL}    & ECCV'22& INet-RN50
& 82.5  & 91.0  & 93.9   
& 38.7  & 49.1  & 55.4  
& 59.1  & 71.2  & 76.3    \\ \shline
OTAM (Baseline)~\cite{OTAM}    & CVPR'20 & CLIP-RN50
& 85.4  & 92.5  & 94.5   
& 39.9  & 48.2  & 51.9  
& 63.0  & 71.2  & 76.6    \\
CLIP-Freeze~\cite{CLIP}    & ICML'21 & CLIP-RN50
& 84.6  & 92.9  & 94.5   
& 26.8  & 33.7  & 36.3  
& 51.4  & 65.2  & 71.0    \\
\rowcolor{Gray}
\textbf{CLIP-FSAR}    & -& CLIP-RN50
& 91.3  & 95.1  & 97.0   
& 52.0  & 54.0  & 55.8  
& 69.2  & 77.6  & 80.3    \\
\rowcolor{Gray}
\rowcolor{Gray}
\textcolor{liGray}{\textbf{CLIP-FSAR$^{\dagger}$}}    & \textcolor{liGray}{-} & \textcolor{liGray}{CLIP-RN50}
& \textcolor{liGray}{92.4}  & \textcolor{liGray}{95.4}  & \textcolor{liGray}{97.0}   
& \textcolor{liGray}{52.1}  & \textcolor{liGray}{54.0}  & \textcolor{liGray}{55.8}  
& \textcolor{liGray}{69.4}  & \textcolor{liGray}{78.3}  & \textcolor{liGray}{80.7}    \\
\shline
OTAM (Baseline)~\cite{OTAM}    & CVPR'20 & CLIP-ViT-B
& 95.8  & 98.2  & 98.8   
& 43.3  & 53.5  & 57.5   
& 72.5  & 81.6  & 83.9     \\
CLIP-Freeze~\cite{CLIP}    & ICML'21 & CLIP-ViT-B
& 89.7  & 94.3  & 95.7   
& 29.5  & 38.0  & 42.5 
& 58.2  & 72.7  & 77.0     \\
\rowcolor{Gray}
\textbf{CLIP-FSAR}     & -& CLIP-ViT-B
& \textbf{96.6}  & \textbf{98.4}  & \textbf{99.0}    
& \textbf{54.5}  & \textbf{58.6}  & \textbf{61.8}   
& \textbf{75.8}  & \textbf{84.1}  & \textbf{87.7}    \\
\rowcolor{Gray}
\rowcolor{Gray}
\textcolor{liGray}{\textbf{CLIP-FSAR$^{\dagger}$}}     & \textcolor{liGray}{-}& \textcolor{liGray}{CLIP-ViT-B}
& \textcolor{liGray}{97.0}  & \textcolor{liGray}{98.5}  & \textcolor{liGray}{99.1}    
& \textcolor{liGray}{54.6}  & \textcolor{liGray}{59.4}  & \textcolor{liGray}{61.8}  
& \textcolor{liGray}{77.1}  & \textcolor{liGray}{84.1}  & \textcolor{liGray}{87.7}     \\
%
%
%
\end{tabular}

\label{tab:compare_SOTA_2}
\vspace{-1mm}
\end{table*}

\noindent \textbf{Evaluation protocol.}
Following prior works~\cite{OTAM,HyRSM,TRX}, we report the performance of our CLIP-FSAR under the 5-way $K$-shot setting with $K$ ranges from 1 to 5. 

\noindent \textbf{Implementation details.}
%
Our experimental implementations are developed on the PyTorch~\cite{paszke2019pytorch} library and trained by Adam~\cite{adam} optimizer.
For simplicity, we adopt OTAM~\cite{OTAM} as the baseline alignment metric unless otherwise stated.
%
Following previous methods~\cite{OTAM,HyRSM}, 8 frames are uniformly and sparsely sampled from each video to encode video representation. 
During the training process, several standard data augmentation techniques, such as random cropping and color jitters, are employed.
Two CLIP models are considered to validate our CLIP-FSAR, namely CLIP-RN50 (ResNet-50)~\cite{Resnet} and CLIP-ViT-B (ViT-B/16)~\cite{ViT}.
In many-shot scenarios (\eg, 5-shot), we adopt the simple but effective average principle~\cite{prototypical} to generate the mean support features before inputting to the prototype modulation.
For inference, the average accuracy of 10,000 randomly sampled few-shot tasks is reported in our experiments.
%
%
%

%
%
\noindent \textbf{Baselines.}
Existing few-shot action recognition methods such as OTAM~\cite{OTAM}, TRX~\cite{TRX}, ITANet~\cite{ITANet}, and HyRSM~\cite{HyRSM} all use the single-modal ImageNet pre-training. 
To verify the effectiveness of our approach, we mainly consider two typical baseline methods with CLIP initialization. 
The first is OTAM, where we replace its original ImageNet pre-training with the visual encoder of CLIP for end-to-end training. 
The second is CLIP-Freeze, which directly leverages the features output from the pre-trained visual encoder and performs  few-shot matching using the OTAM metric without re-training on the base dataset.
\begin{table}[ht]
\centering
\footnotesize
\tablestyle{5pt}{1.2}
\caption{%
Generalization experiment. 
We plug our approach into several existing few-shot action recognition methods.
}
\vspace{-3mm}
\resizebox{0.47\textwidth}{24mm}{
\begin{tabular}
{l|c|cc|cc}
			
  \multirow{2}{*}{Method}  & \multicolumn{1}{c|}{\multirow{2}{*}{\hspace{-2mm} Pre-training \hspace{-2mm}}} 
& \multicolumn{2}{c|}{{SSv2-Small}}  & \multicolumn{2}{c}{{Kinetics}}  \\
& \multicolumn{1}{c|}{}  
& \multicolumn{1}{l}{1-shot}  & 5-shot 
&  \multicolumn{1}{l}{1-shot}   & 5-shot  \\ \shline

Bi-MHM~\cite{HyRSM} & INet-RN50  & 38.0  & 48.9  & 72.3   & 84.5    \\
Bi-MHM~\cite{HyRSM} & CLIP-RN50  & 40.6  & 52.2  & 77.1   & 88.9    \\
\rowcolor{Gray}
 \textbf{Ours (Bi-MHM)} & CLIP-RN50   & \textbf{52.3}  & \textbf{55.6}  & \textbf{87.7} &  \textbf{92.1} \\ 
\shline
ITANet~\cite{ITANet} & INet-RN50  & 39.8  & 53.7  &  73.6   & 84.3    \\
ITANet~\cite{ITANet} & CLIP-RN50  & 40.2 & 56.5  & 78.5   & 89.0   \\
\rowcolor{Gray}
 \textbf{Ours (ITANet)} & CLIP-RN50   & \textbf{49.6}  & \textbf{58.2}  & \textbf{79.3} &  \textbf{89.6} \\
 \shline
TRX~\cite{TRX} & INet-RN50  & 36.0  & 56.7  & 63.6   & 85.9    \\
TRX~\cite{TRX} & CLIP-RN50  & 39.4 & 56.8  & 71.8   & 89.2    \\
\rowcolor{Gray}
 \textbf{Ours (TRX)} & CLIP-RN50   & \textbf{51.5}  & \textbf{57.1}  & \textbf{86.7} &  \textbf{92.2} \\


%
%
\end{tabular}
}
\label{tab:ablation_extend_metric}
\vspace{-1mm}
\end{table}

\begin{table}[ht]
\centering
\footnotesize
\tablestyle{5pt}{1.2}
\caption{
Ablation study of each component in our CLIP-FSAR.
}
\vspace{-3mm}
\setlength{
    \tabcolsep}{
    1.4mm}{
\begin{tabular}
{cc|cc|cc}
			
  \multirow{2}{*}{Video-text}  & \multicolumn{1}{c|}{\multirow{2}{*}{Prototype modulation}} 
& \multicolumn{2}{c|}{{SSv2-Small}}  & \multicolumn{2}{c}{{Kinetics}}  \\
& \multicolumn{1}{c|}{}  
& \multicolumn{1}{l}{1-shot}  & 5-shot 
&  \multicolumn{1}{l}{1-shot}   & 5-shot  \\ \shline

 &   & 39.9  & 51.9   & 76.4   & 88.9    \\
\checkmark &   & 41.4  & 55.5   & 80.4   & 91.6     \\ 
 & \checkmark & 50.8  & 55.3   & 86.5   & 91.3     \\ 
\checkmark &  \checkmark & \textbf{52.0}  & \textbf{55.8}   & \textbf{87.6}   & \textbf{91.9}     \\ 
%
%
\end{tabular}
}
\label{tab:ablation_1}
\vspace{-1mm}
\end{table}

\subsection{Comparison with state-of-the-art methods}

To verify the effectiveness of the proposed framework, 
we compare the performance of our CLIP-FSAR over current state-of-the-art few-shot action recognition methods on five standard datasets.
%
The results are summarized in Table~\ref{tab:compare_SOTA_1} and Table~\ref{tab:compare_SOTA_2}.
From the experimental results, we can make the following observations: 
(a) Compared with the OTAM baseline, our method can significantly boost performance by leveraging the multi-modal knowledge of CLIP. For example, under the 5-way 1-shot SSv2-Full setting, our method achieves 14.0\% and 11.7\% gains using CLIP-RN50 and CLIP-ViT-B, respectively.
Notably, our CLIP-FSAR based on CLIP-ViT-B consistently outperforms other state-of-the-art techniques, demonstrating the effectiveness of our method.  
(b) By comparing the results of OTAM and CLIP-Freeze with the same CLIP's visual encoder, we can see that re-training CLIP on the few-shot action recognition task can make it adapt to the downstream task and further improve the performance.
(c) CLIP-FSAR based on CLIP-VIT-B generally achieves results superior to that based on CLIP-RN50, indicating that a stronger pre-trained model leads to better few-shot generalization.
Moreover, multimodal pre-trained CLIP displays better performance than the ImageNet pre-training counterpart.
(d) 
The performance margin between CLIP-FSAR and baseline is more significant with a smaller shot and slowly shrinks with the shot increasing.
We attribute this to the fact that introducing text semantic cues 
would be more effective when visual information is insufficient.
Similarly, compared with baselines, the performance gain based on CLIP-RN50 is more considerable than CLIP-ViT-B.
On the Kinetics dataset, the 1-shot performance margin is 11.2\%  (87.6\% vs. 76.4\%)  on CLIP-RN50, while 1.5\% (89.7\% vs. 88.2\%) on CLIP-VIT-B. 
(e) By incorporating the results of video-text matching into the few-shot classification, \ie, CLIP-FSAR$^{\dagger}$, the performance is also improved to some extent.

%
%
%
To further verify the pluggability of our framework, we apply the proposed method as a plug-and-play component to the existing metrics or methods, such as Bi-MHM~\cite{HyRSM}, ITANet~\cite{ITANet}, and TRX~\cite{TRX}. 
From Table~\ref{tab:ablation_extend_metric},
we can find that a notable performance improvement is also achieved when extending our CLIP-FSAR to these techniques, suggesting that our approach is a generic architecture.

\subsection{Ablation study}

To study the contribution of each component in CLIP-FSAR, we
conduct a series of detailed ablation studies.
Unless otherwise stated, we adopt the CLIP-RN50 model as the default setting for comparative experiments.

\vspace{1mm}
\noindent \textbf{Component analysis.}
In Table~\ref{tab:ablation_1}, we investigate the role of the video-text contrastive objective and prototype modulation in the proposed CLIP-FSAR.
By performing video-text contrastive, we obtain 4.0\% and 2.7\% 1-shot and 5-shot improvements on Kinetics, respectively.
This consistent promotion indicates the importance of adapting CLIP to the few-shot video task.
Likewise, implementing prototype modulation provides 10.9\% and 3.4\% performance gains over the baseline on the SSv2-Small dataset.
We observe that the performance gain is particularly
significant in the 1-shot scenario, suggesting that it would be more effective to supplement textual semantic information when visual information is very limited.
Besides, the best performance is achieved by training both components jointly, which empirically validates that the video-text contrastive objective and prototype modulation complement each other.

\begin{table}[t]
\centering
\footnotesize
\tablestyle{5pt}{1.2}
\caption{
Comparison results of different number of temporal Transformer layers on the SSv2-Small and Kinetics datasets.
}
\vspace{-3mm}
\begin{tabular}
{l|cc|cc}
			
  \multirow{2}{*}{Setting}  
& \multicolumn{2}{c|}{{SSv2-Small}}  & \multicolumn{2}{c}{{Kinetics}}  \\
 \multicolumn{1}{c|}{}  
& \multicolumn{1}{l}{1-shot}  & 5-shot 
&  \multicolumn{1}{l}{1-shot}   & 5-shot  \\ \shline

Temporal Transformer $\times 1$  & 52.0  & 55.8   & \textbf{87.6}   & \textbf{91.9}    \\
Temporal Transformer $\times 2$  & 52.1  & 56.4   & 87.4   & 91.4    \\
Temporal Transformer $\times 3$  & \textbf{52.4}  & \textbf{56.6}   & 86.7   & 91.2    \\
Temporal Transformer $\times 4$  & 51.6  & 55.8   & 85.8   & 90.9    \\
Temporal Transformer $\times 5$  & 50.8  & 55.1   & 85.2   & 90.4    \\
%
%
\end{tabular}
\label{tab:ablation_transformer}
\end{table}

\begin{table}[t]
\centering
\vspace{+2mm}
\footnotesize
\tablestyle{5pt}{1.2}
\caption{Comparison of different prototype modulation schemes.
``Temporal Transformer-only" indicates that the temporal Transformer is used to model the temporal relations without incorporating text features for prototype modulation.
}
\vspace{-3mm}
\begin{tabular}
{l|cc|cc}
			
  \multirow{2}{*}{Setting}  
& \multicolumn{2}{c|}{{SSv2-Small}}  & \multicolumn{2}{c}{{Kinetics}}  \\
 \multicolumn{1}{c|}{}  
& \multicolumn{1}{l}{1-shot}  & 5-shot 
&  \multicolumn{1}{l}{1-shot}   & 5-shot  \\ \shline
Temporal Transformer-only  & 40.5  & 52.3   & 77.9   & 90.0    \\
\shline
Bi-LTSM  & 45.2  & 55.3   & 80.0   & 89.1    \\
Bi-GRU  & 47.1  & 55.5   & 80.5   & 89.5    \\
Temporal Transformer (Ours) & \textbf{52.0}  & \textbf{55.8}   & \textbf{87.6}   & \textbf{91.9}   \\

%
%
\end{tabular}
\label{tab:ablation_temporal_manner}
\vspace{-2mm}
\end{table}

\begin{table}[t]
\small
\vspace{0pt}
\vspace{+2mm}
\centering
\tablestyle{5pt}{1.2}
\caption{
 Comparison of shared weights and non-shared weights.
}
\vspace{-3mm}
\setlength{
    \tabcolsep}{
    3mm}{
    \begin{tabular}
{l|cc|cc}
			
  \multirow{2}{*}{Setting}  
& \multicolumn{2}{c|}{{SSv2-Small}}  & \multicolumn{2}{c}{{Kinetics}}  \\
 \multicolumn{1}{c|}{}  
& \multicolumn{1}{l}{1-shot}  & 5-shot 
&  \multicolumn{1}{l}{1-shot}   & 5-shot  \\ \shline

Non-shared  & 51.1  & 55.0  & 79.5   & 84.7    \\
Shared (Ours)  & \textbf{52.0}  & \textbf{55.8}   & \textbf{87.6}   & \textbf{91.9}   \\

%
%
\end{tabular}
}
\label{tab:ablation_shared_transformer}
\end{table}
\vspace{1mm}
\noindent \textbf{Changing the number of temporal Transformer layers.}
To explore
the influence of the number of applied temporal Transformer layers, we
conduct ablation studies 
on SSv2-Small and Kinetics.
The results are presented in Table~\ref{tab:ablation_transformer}, we can notice that on SSv2-Small, the performance slowly improves as the number of Transformer layers increases, and overfitting starts to appear beyond a specific value.
Differently, a single Transformer layer is sufficient for optimal performance on the Kinetics dataset, which we attribute to the fact that this dataset is relatively appearance-based and temporal modeling is less critical for recognition~\cite{chen2021deep}. 
In order to balance precision and efficiency, we adopt a single temporal Transformer layer as our default setting.
%
%
%

\vspace{1mm}
\noindent \textbf{Influence of the prototype modulation scheme.}
In Table~\ref{tab:ablation_temporal_manner}, 
we conduct ablation experiments where
several commonly used temporal aggregation operators like Bi-LSTM~\cite{BiLSTM}, Bi-GRU~\cite{BiGRU}, and temporal Transformer~\cite{Transformer} are compared to further analyze the effect of different 
prototype modulation schemes.
We observe that the temporal Transformer consistently outperforms other competitors due to flexible information interaction and the powerful generalization capability of Transformer~\cite{transformer-xl,HyRSM}.
In addition, the property that the output of the Transformer is independent with the relative position of the input semantic features and visual features in the prototype modulation mechanism makes it more suitable for our framework.
To further analyze whether the temporal Transformer applied to support and query videos should share weights, we conduct an experimental comparison in Table~\ref{tab:ablation_shared_transformer}.
The experimental results reveal that sharing weights allows query and support samples to be in the same feature space and thus achieve better few-shot action recognition results.

\vspace{-1mm}
\begin{figure}[t]
  \centering
   \includegraphics[width=0.99\linewidth]{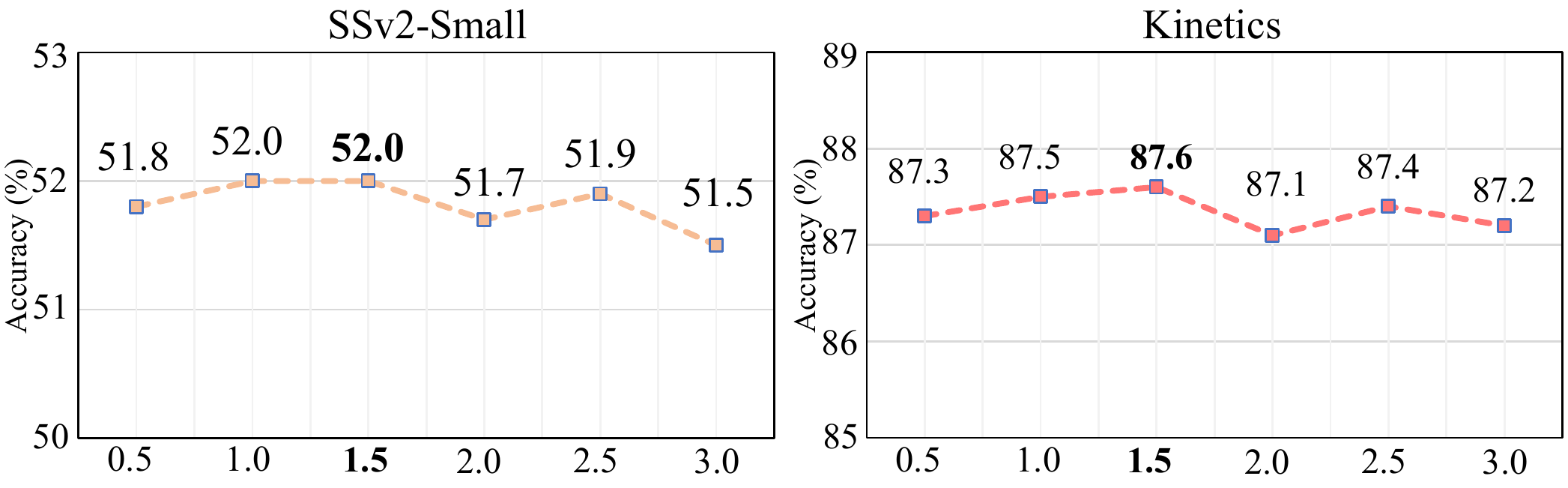}
\vspace{-2mm}
   \caption{
   Sensitivity analysis of $\alpha $ on SSv2-Small and Kinetics.
   }
   \label{fig:Hyperparameter}
   \vspace{-0mm}
\end{figure}

\begin{figure}[t]
  \centering
   \includegraphics[width=0.95\linewidth]{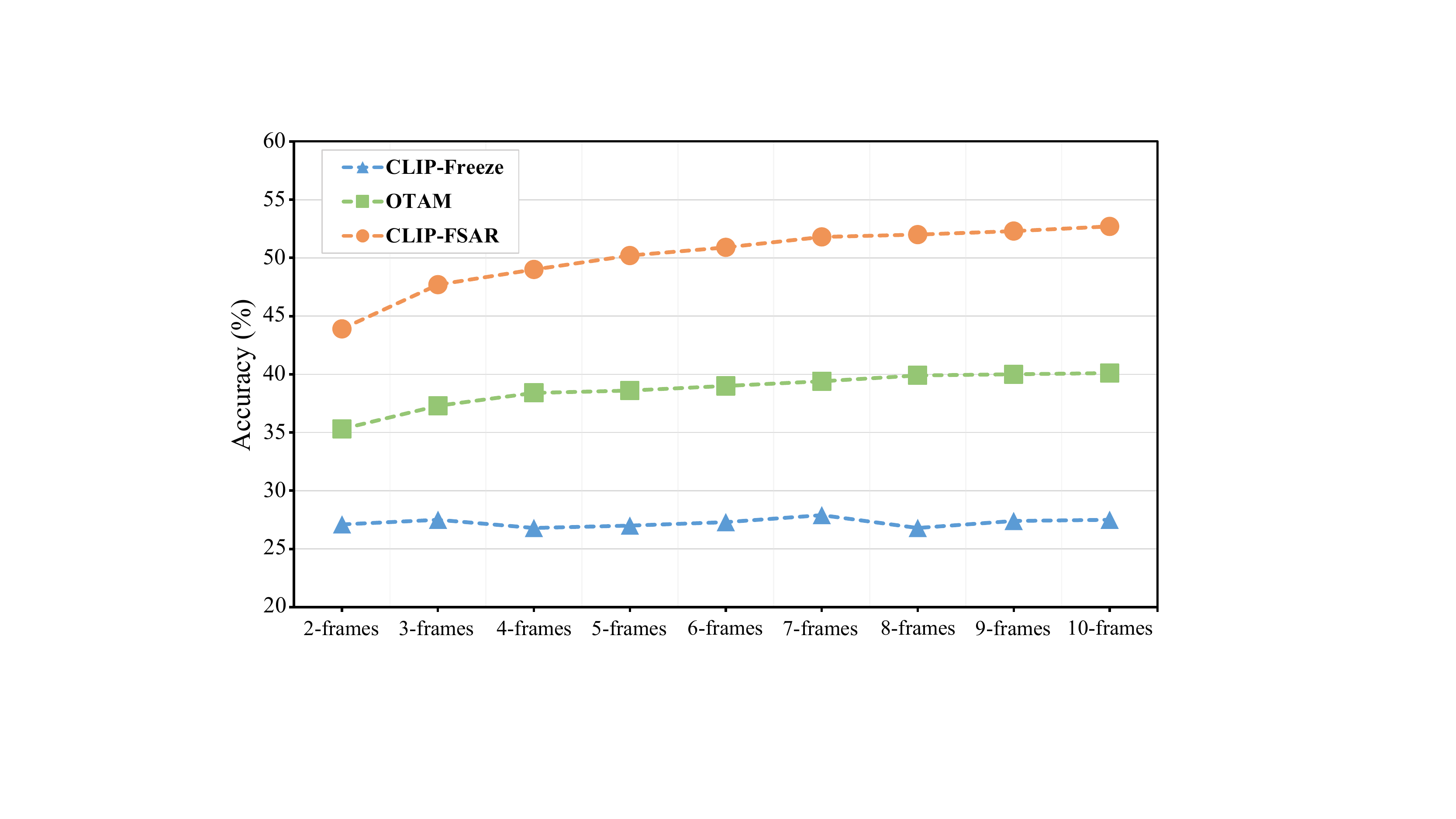}
\vspace{-2mm}
   \caption{
   Performance comparison with different numbers
of input video frames under the 5-way 1-shot setting on SSv2-Small.
}
   \label{fig:frame_ablation}
\end{figure}


\vspace{1mm}
\noindent \textbf{Hyperparameter sensitivity analysis.}
In Equation~\ref{vt-eq}, we introduce a hyperparameter $\alpha$ to balance two loss terms.
As shown in Figure~\ref{fig:Hyperparameter}, we report the ablation results.
We can observe that the impact of this parameter on performance is relatively small, and CLIP-FSAR always maintains the leading performance.
%

\vspace{1mm}
\noindent \textbf{Varying the number of input video frames.}
To make a fair comparison with previous methods~\cite{OTAM,TRX,ITANet}, we sample 8 frames from each video to encode video representation in experiments.
We further analyze the effect of different numbers of input video frames in Figure~\ref{fig:frame_ablation}.
The results demonstrate that as the number of input video frames increases, the performance of CLIP-FSAR slowly increases and eventually saturates. 
Remarkably, our method consistently outperforms the baselines, indicating the scalability of CLIP-FSAR.
An interesting phenomenon is the oscillation result of CLIP-Freeze, which we attribute to the fact that CLIP is  initially pre-trained for image fields and can not perceive motion patterns in videos without adaptation.
%

\begin{figure}[t]
  \centering
   \includegraphics[width=0.98\linewidth]{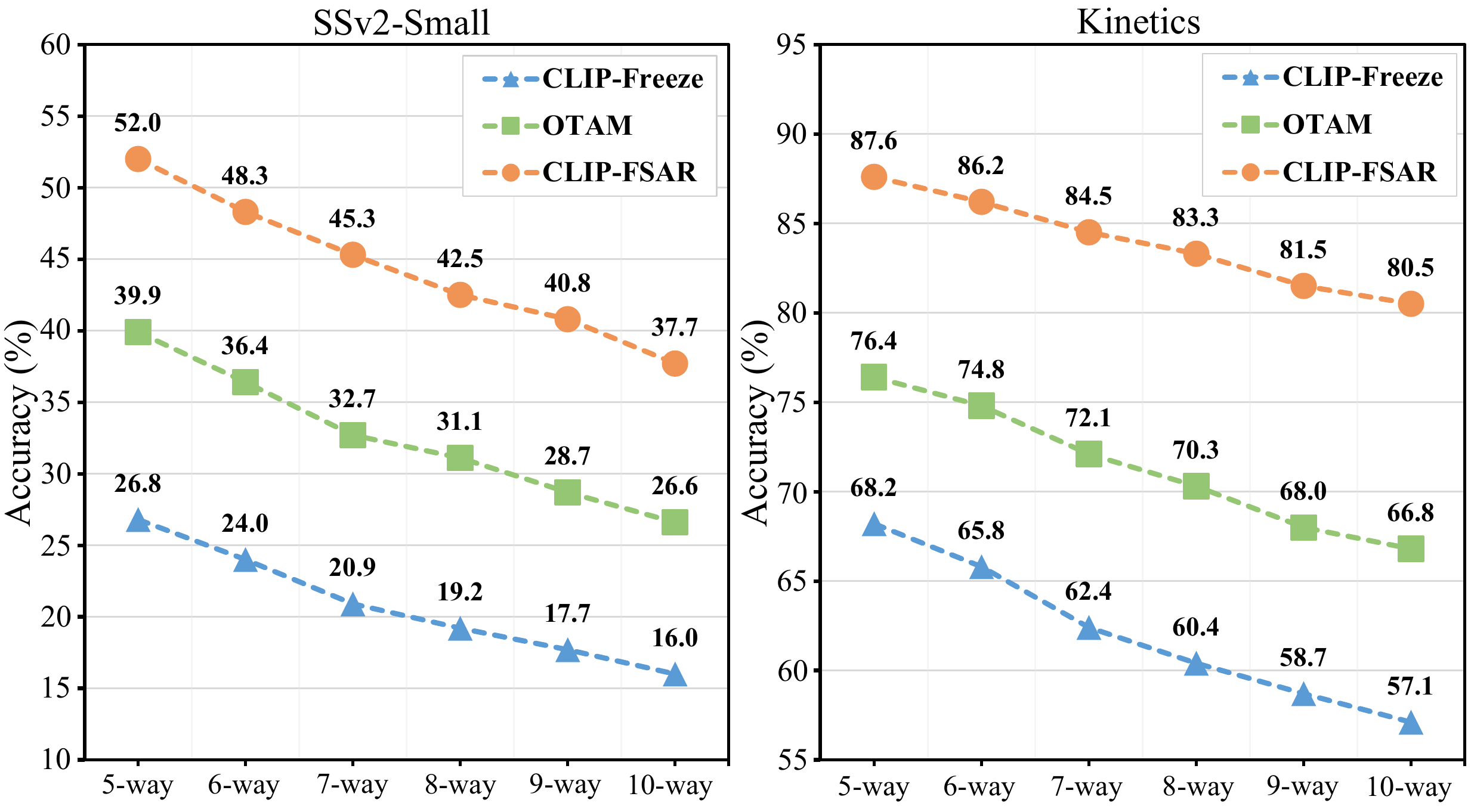}
\vspace{-2mm}
   \caption{
   N-way 1-shot results of our CLIP=FSAction and
other baseline methods with N varying from 5 to 10.
   }
   \label{fig:n_way_ablation}
\end{figure}

\vspace{1mm}
\noindent \textbf{$N$-way classification.}
We also ablate the effect of varying $N$ on the few-shot performance. 
The  $N$-way 1-shot comparison results are depicted in Figure~\ref{fig:n_way_ablation}, where $N$ varies from 5 to 10. 
We can observe that a larger $N$ represents a greater difficulty in classification, and the performance is lower. 
For example, the 10-way SSv2-Small result of CLIP-FSAR decreases by 14.3\% compared to the 5-way result.
Nevertheless, CLIP-FSAR still significantly outperforms the comparison methods under various settings, indicating the effectiveness of our method.

%
\begin{table}[t]
\centering
\footnotesize
\tablestyle{5pt}{1.2}
\caption{
Empirical analysis of replacing the visual encoder of our CLIP-FSAR with several ImageNet pre-trained backbones.
}
\vspace{-3mm}
\begin{tabular}
{l|c|cc|cc}
			
  \multirow{2}{*}{Method}  & \multicolumn{1}{c|}{\multirow{2}{*}{Pre-training}} 
& \multicolumn{2}{c|}{{SSv2-Small}}  & \multicolumn{2}{c}{{Kinetics}}  \\
& \multicolumn{1}{c|}{}  
& \multicolumn{1}{l}{1-shot}  & 5-shot 
&  \multicolumn{1}{l}{1-shot}   & 5-shot  \\ \shline

OTAM & INet-RN18  & 35.8  & 47.1  & 68.6   & 78.3    \\
\rowcolor{Gray}
 CLIP-FSAR & INet-RN18   & \textbf{46.0}  & \textbf{52.4}  & \textbf{72.9} &  \textbf{84.1} \\ 
\shline
 OTAM  & INet-RN34  & 36.2  & 47.9  & 71.0   & 80.4     \\
 \rowcolor{Gray}
CLIP-FSAR  & INet-RN34   & \textbf{46.5}  & \textbf{53.0}  & \textbf{73.9}   & \textbf{85.3}    \\ 
\shline
 OTAM  & INet-RN50  & 36.4  & 48.0  & 72.3   & 84.2    \\
 \rowcolor{Gray}
 CLIP-FSAR  & INet-RN50   & \textbf{46.7}  & \textbf{53.3}  & \textbf{75.5}   & \textbf{85.9}    \\ 

%
%
\end{tabular}
\label{tab:imagenet}
\vspace{-1mm}
\end{table}

\vspace{1mm}
\noindent \textbf{Generalization analysis.}
Our CLIP-FSAR is based on the pre-trained visual and text encoder of CLIP for comparison experiments. 
To further explore whether our framework can be extended to the standard single-modal pre-trained initialization, we replace the visual encoder with the models pre-trained on ImageNet and keep the text encoder unchanged.
As illustrated in Table~\ref{tab:imagenet}, we conduct experiments on three ImageNet-initialized models of different depths, ResNet-18, ResNet-34, and ResNet-50, respectively. 
The results adequately suggest that our method can effectively improve the performance of these baseline methods, \eg, CLIP-FSAR boosts the 1-shot SSv2-Small performance from 35.8\% to 46.0\% using ResNet-18, which fully demonstrates the generalization of our framework.
In Table~\ref{tab:ablation_extend_2}, we also apply our CLIP-FSAR framework to a self-supervised DINO~\cite{zhang2022dino} method. 
Concretely, we replace the visual encoder in CLIP-FSAR with an open-available DINO ResNet-50 model that has been self-supervised pre-trained on ImageNet and keeps the text encoder unchanged.
From the results, we notice that based on OTAM and TRX, our framework still achieves impressive performance gains compared to the naive baselines, indicating the strong generality of our method.
%

\vspace{1mm}
\noindent \textbf{Zero-shot performance.}
Although our CLIP-FSAR is originally designed for the few-shot action recognition task, we can also perform zero-shot classification using the video-text contrastive objective. 
For convenience, we utilize the same data splits as the few-shot evaluation, training on the training set and performing zero-shot classification on the test set.
The 5-way zero-shot experimental comparisons are shown in Table~\ref{tab:ablation_zero-shot}. 
It can be found that our method shows better performance compared to CLIP-Freeze.
For instance, our method achieves 11.2\% performance improvement over CLIP-Freeze on the HMDB dataset with CLIP-ViT-B.
%
This can be explained by the adaptation of our CLIP-FSAR to the video task.

\begin{table}[t]
\centering
\footnotesize
\tablestyle{5pt}{1.2}
\caption{%
Generalizability experiments of migrating our CLIP-FSAR framework to a self-supervised DINO visual encoder.
}
\vspace{-3mm}
\begin{tabular}
{l|c|cc|cc}
  \multirow{2}{*}{Method}  & \multicolumn{1}{c|}{\multirow{2}{*}{\hspace{-2mm} Pre-training \hspace{-2mm}}} 
& \multicolumn{2}{c|}{{SSv2-Small}}  & \multicolumn{2}{c}{{Kinetics}}  \\
& \multicolumn{1}{c|}{}  
& \multicolumn{1}{l}{1-shot}  & 5-shot 
&  \multicolumn{1}{l}{1-shot}   & 5-shot  \\ \shline

OTAM & DINO-RN50  & 39.8  & 52.5  & 66.3   & 77.5    \\
\rowcolor{Gray}
 \textbf{Ours (OTAM)} &  DINO-RN50   & \textbf{47.3}  & \textbf{53.4}  & \textbf{73.3} &  \textbf{84.6} \\ 
\shline
TRX &  DINO-RN50  & 38.2  & 57.5  &  58.7   & 78.4    \\
\rowcolor{Gray}
 \textbf{Ours (TRX)} &  DINO-RN50  & \textbf{44.8}  & \textbf{58.9}  & \textbf{70.2} &  \textbf{85.1} \\
\end{tabular}
\label{tab:ablation_extend_2}
\end{table}

\begin{table}[t]
\centering
\footnotesize
\tablestyle{5pt}{1.2}
\caption{
5-way zero-shot performance comparison with CLIP.
}
\vspace{-3mm}
\begin{tabular}
{l|c|c|c|c}

  Method  & Encoder
& SSv2-Small  & Kinetics  & HMDB51  \\ \shline

CLIP-Freeze & RN-50 & 30.6  & 89.5 & 55.1    \\
 \rowcolor{Gray}
CLIP-FSAR & RN-50 & \textbf{43.3}  & \textbf{89.7}   & \textbf{62.6}  \\
\shline
CLIP-Freeze & ViT-B & 36.0  & 94.2 &  64.4   \\
 \rowcolor{Gray}
CLIP-FSAR & ViT-B & \textbf{46.8}  & \textbf{94.3}   & \textbf{75.6}  \\

%
%
\end{tabular}
\label{tab:ablation_zero-shot}
\vspace{-1mm}
\end{table}

\begin{figure}[t] 
  \centering
    \begin{subfigure}{0.23\textwidth}
      \centering   
      \includegraphics[width=0.99\linewidth]{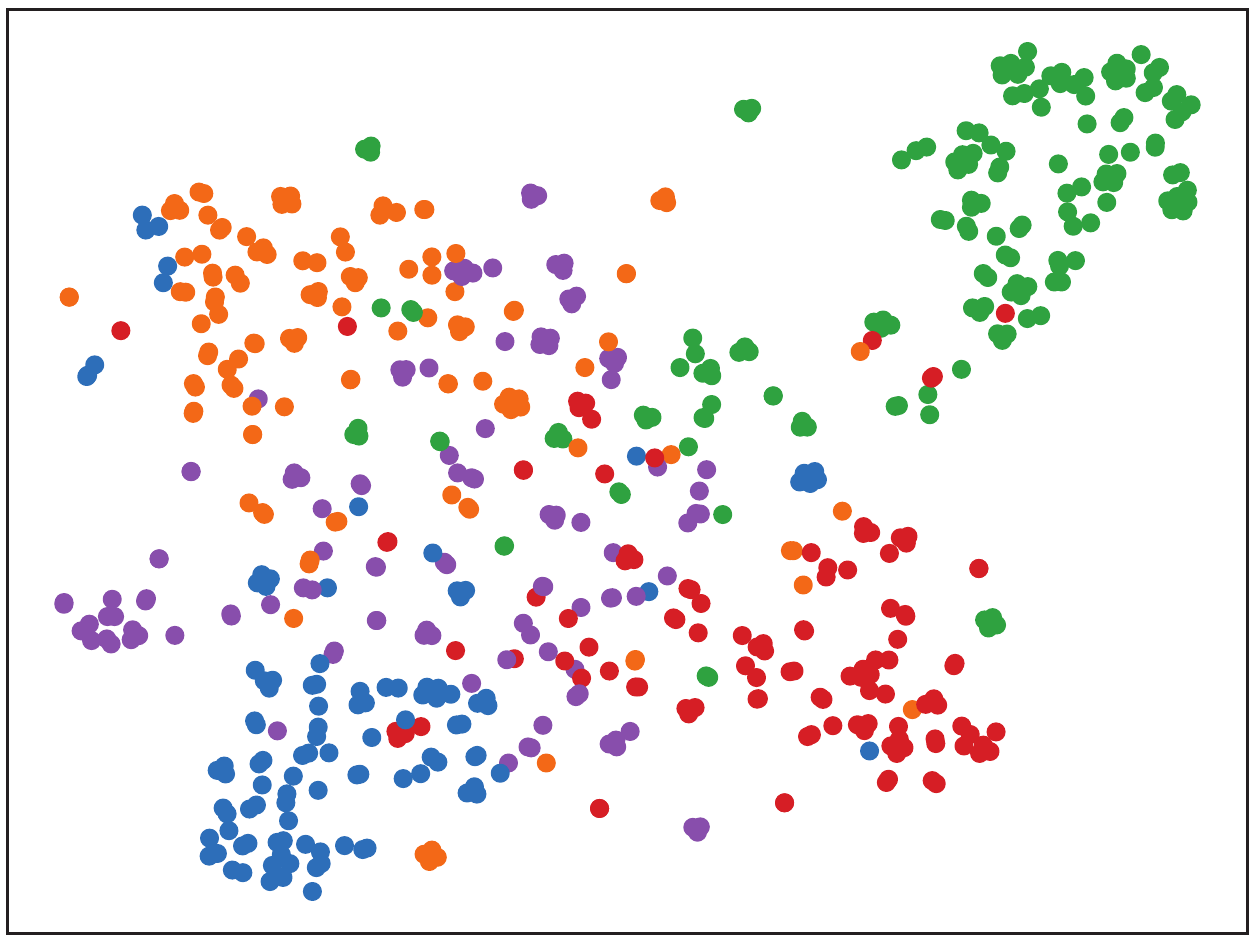}
        \caption{Without prototype modulation}
        \label{fig:sub1}
    \end{subfigure}   
    \begin{subfigure}{0.23\textwidth}
      \centering   
      \includegraphics[width=0.985\linewidth]{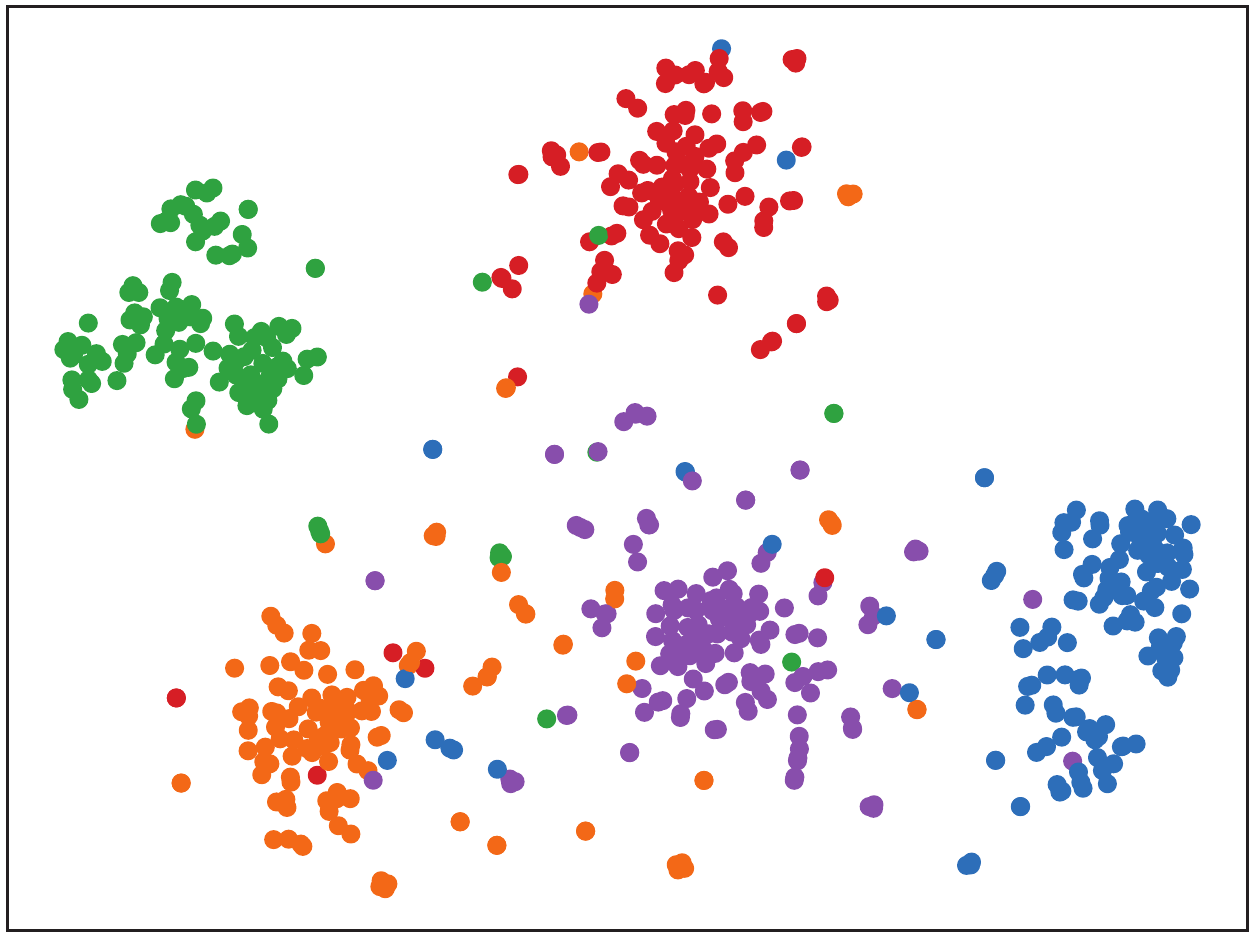}
        \caption{With prototype modulation}
        \label{fig:sub2}
    \end{subfigure}
\vspace{-3mm}
\caption{
\label{fig:tsne}
T-SNE distribution visualization of five action classes on the test set of SSv2-Small. 
Better view in colored PDF.
}
\end{figure}

\begin{figure}[t] 
  \centering
    \begin{subfigure}{0.23\textwidth}
      \centering   
      \includegraphics[width=0.99\linewidth]{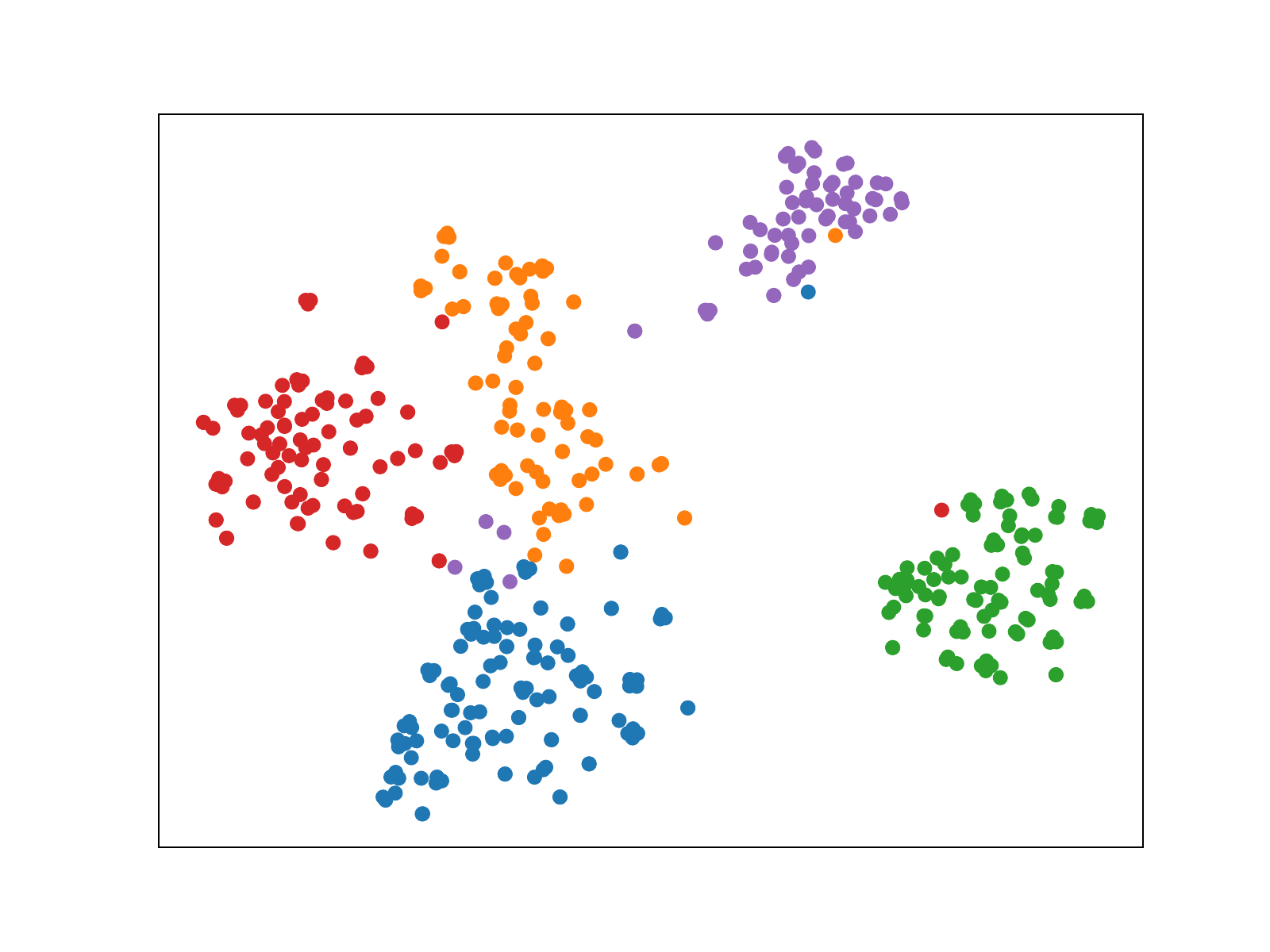}
        \caption{Without prototype modulation}
    \end{subfigure}   
    \begin{subfigure}{0.23\textwidth}
      \centering   
      \includegraphics[width=0.985\linewidth]{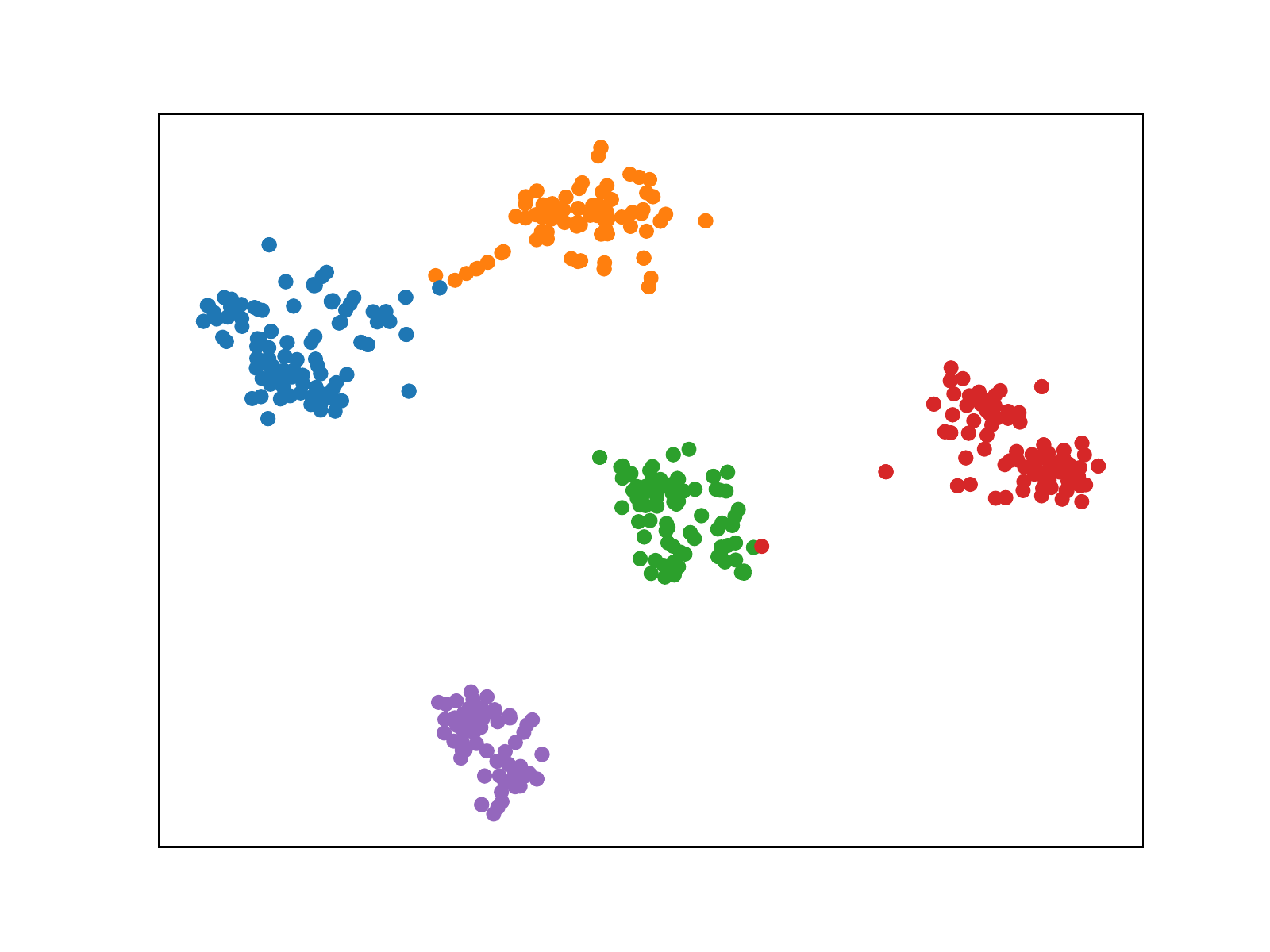}
        \caption{With prototype modulation}
    \end{subfigure}
\vspace{-3mm}
\caption{
\label{fig:tsne_Kinetics}
T-SNE distribution visualization of five action classes on the test set of Kinetics. 
%
The five classes are \textcolor[RGB]{23,118,177}{``playing monopoly"}, \textcolor[RGB]{255,122,4}{``hula hooping"}, \textcolor[RGB]{44,160,44}{``cutting watermelon"}, \textcolor[RGB]{220,45,41}{``shearing sheep"} and \textcolor[RGB]{129,77,177}{``playing drums"}.
}
\end{figure}

\begin{figure}[t] 
  \centering
    \begin{subfigure}{0.23\textwidth}
      \centering   
      \includegraphics[width=0.99\linewidth]{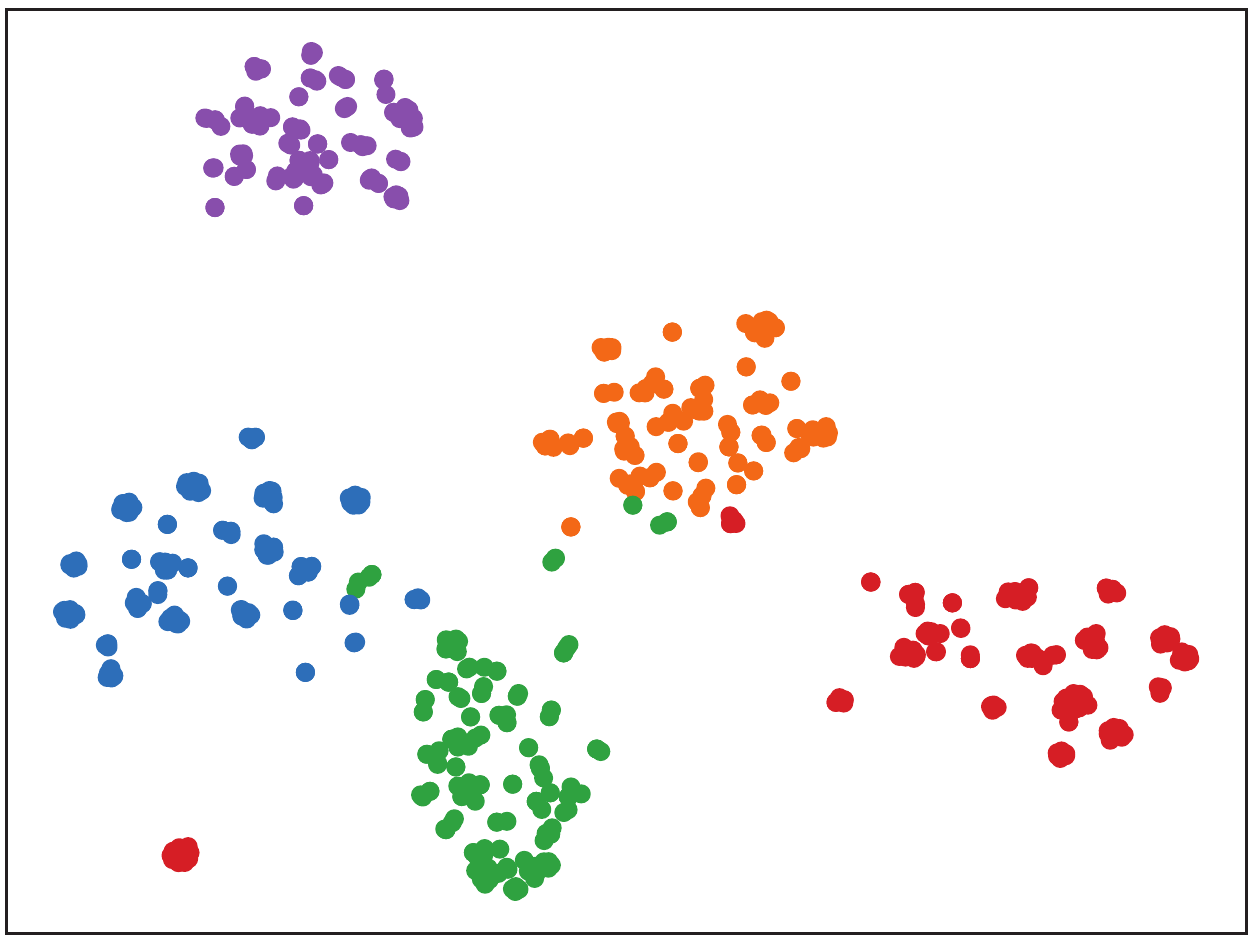}
        \caption{Without prototype modulation}
    \end{subfigure}   
    \begin{subfigure}{0.23\textwidth}
      \centering   
      \includegraphics[width=0.985\linewidth]{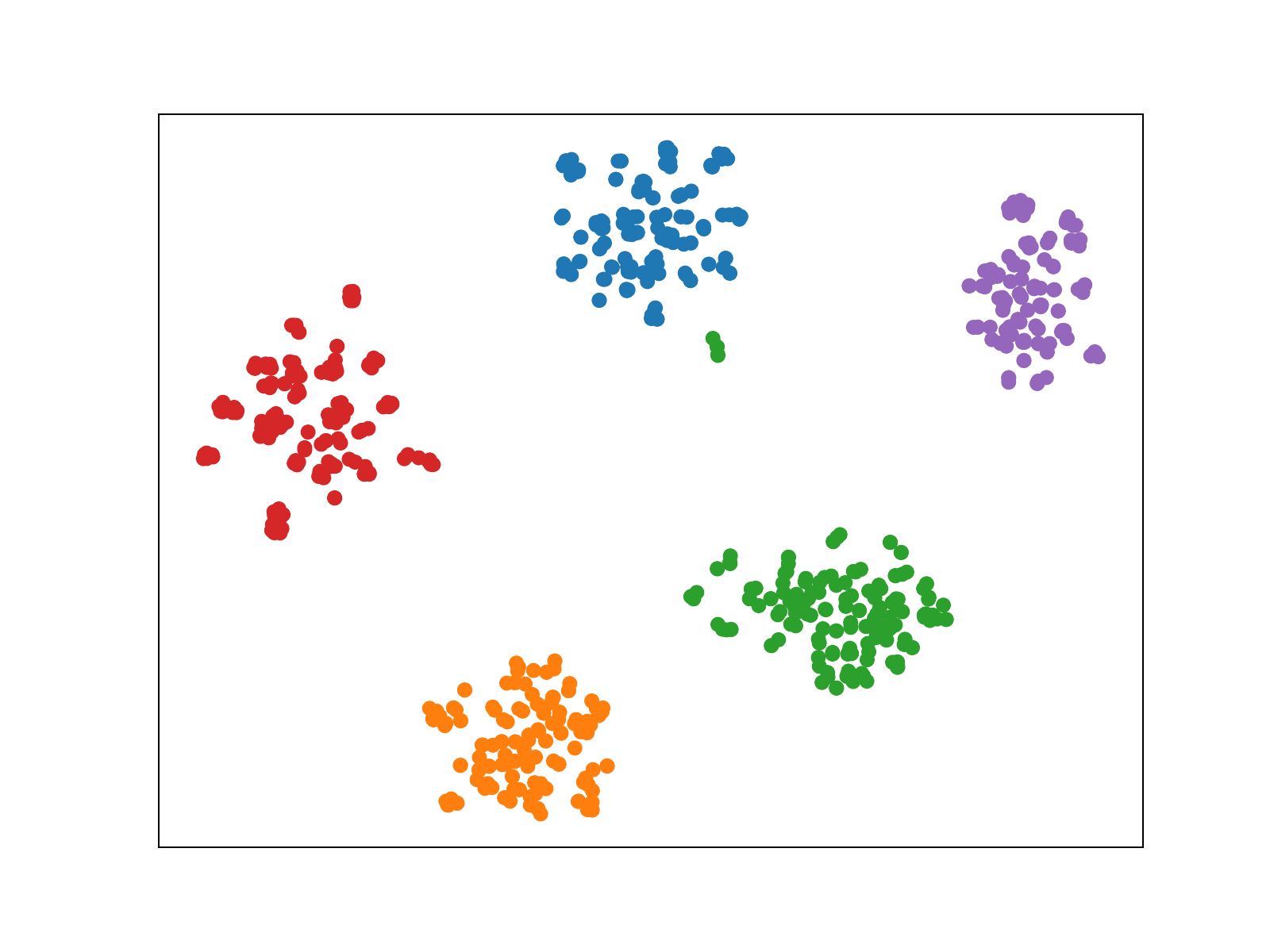}
        \caption{With prototype modulation}
    \end{subfigure}
\vspace{-3mm}
\caption{
\label{fig:tsne_UCF101}
T-SNE distribution visualization of five action classes on the test set of UCF101. 
%
The five classes are \textcolor[RGB]{23,118,177}{``salsa spin"}, \textcolor[RGB]{255,122,4}{``diving"}, \textcolor[RGB]{44,160,44}{``punch"}, \textcolor[RGB]{220,45,41}{``tennis swing"} and \textcolor[RGB]{129,77,177}{``cutting in kitchen"}.
}
\end{figure}

\begin{figure}[t] 
  \centering
    \begin{subfigure}{0.23\textwidth}
      \centering   
      \includegraphics[width=0.99\linewidth]{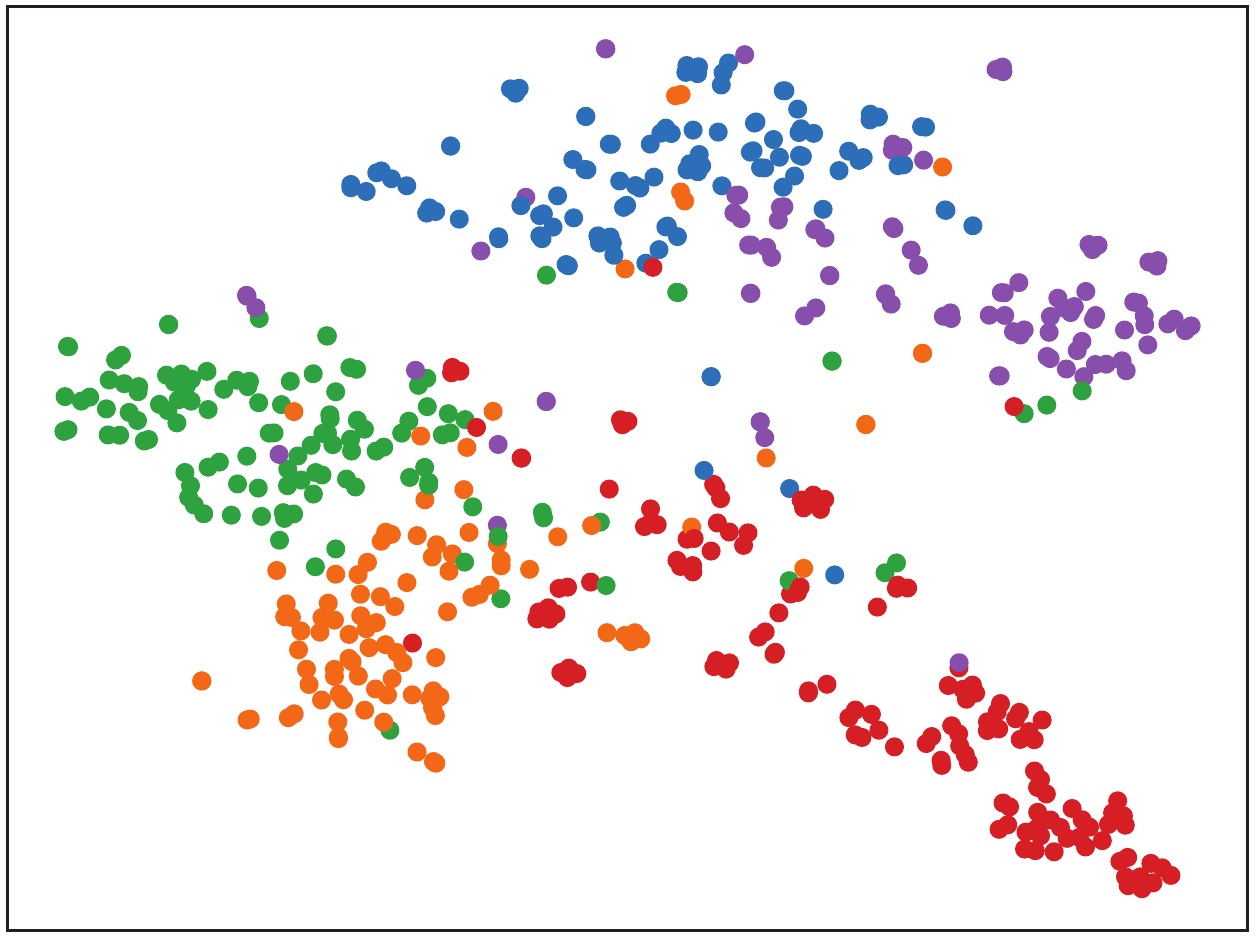}
        \caption{Without prototype modulation}
    \end{subfigure}   
    \begin{subfigure}{0.23\textwidth}
      \centering   
      \includegraphics[width=0.99\linewidth]{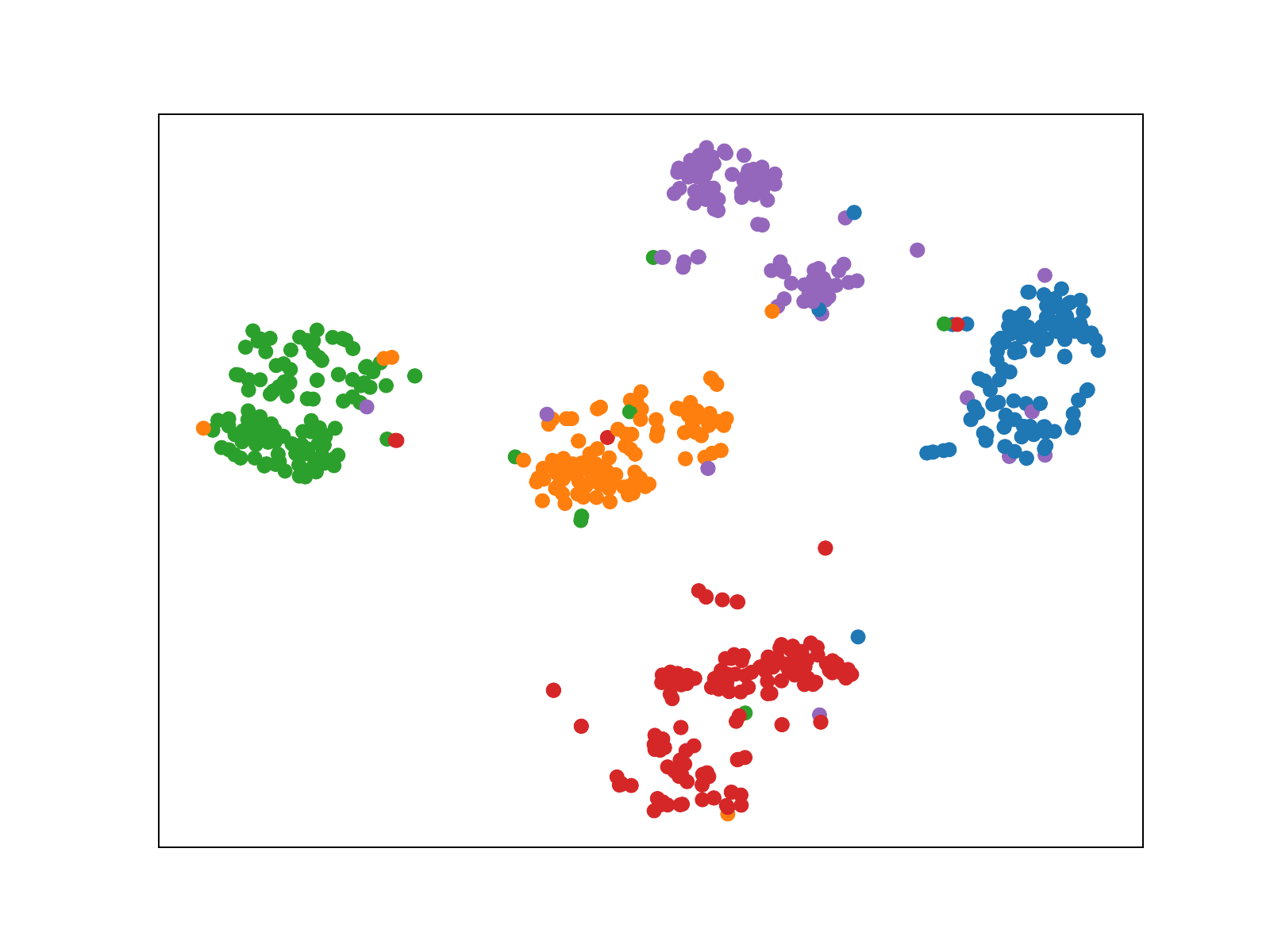}
        \caption{With prototype modulation}
    \end{subfigure}
\vspace{-3mm}
\caption{
\label{fig:tsne_SSv2-full}
T-SNE distribution visualization of five action classes on the test set of SSv2-Full. 
%
The five classes are \textcolor[RGB]{23,118,177}{``Approaching [something] with our camera"}, \textcolor[RGB]{255,122,4}{``Taking [something] out of [something]"}, \textcolor[RGB]{44,160,44}{``Pushing [something] from right to left"}, \textcolor[RGB]{220,45,41}{``Pouring [something] out of [something]"} and \textcolor[RGB]{129,77,177}{``Showing [something] next to [something]."}
}
\end{figure}

\begin{figure}[t] 
  \centering
    \begin{subfigure}{0.23\textwidth}
      \centering   
      \includegraphics[width=0.99\linewidth]{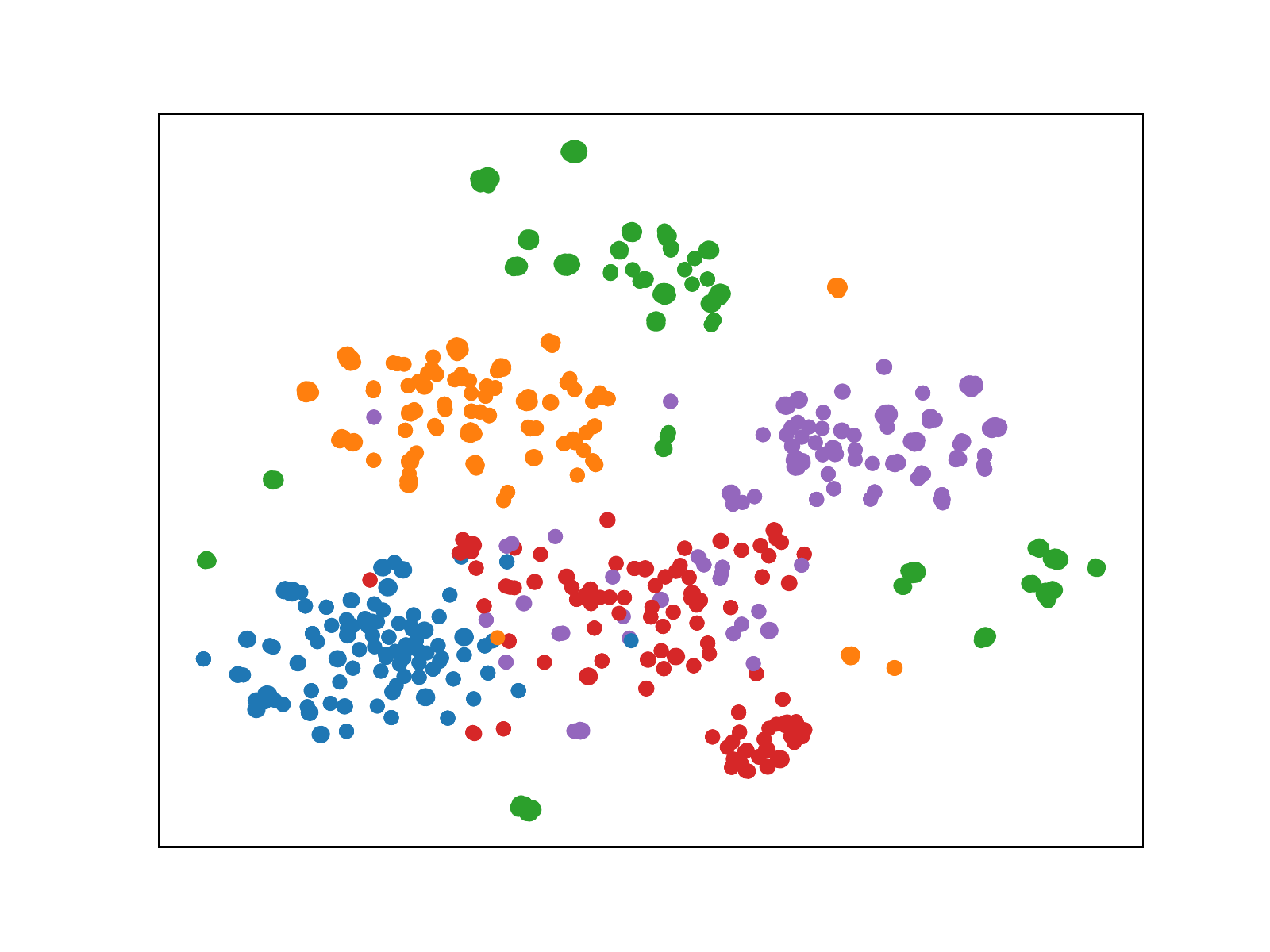}
        \caption{Without prototype modulation}
    \end{subfigure}   
    \begin{subfigure}{0.23\textwidth}
      \centering   
      \includegraphics[width=0.99\linewidth]{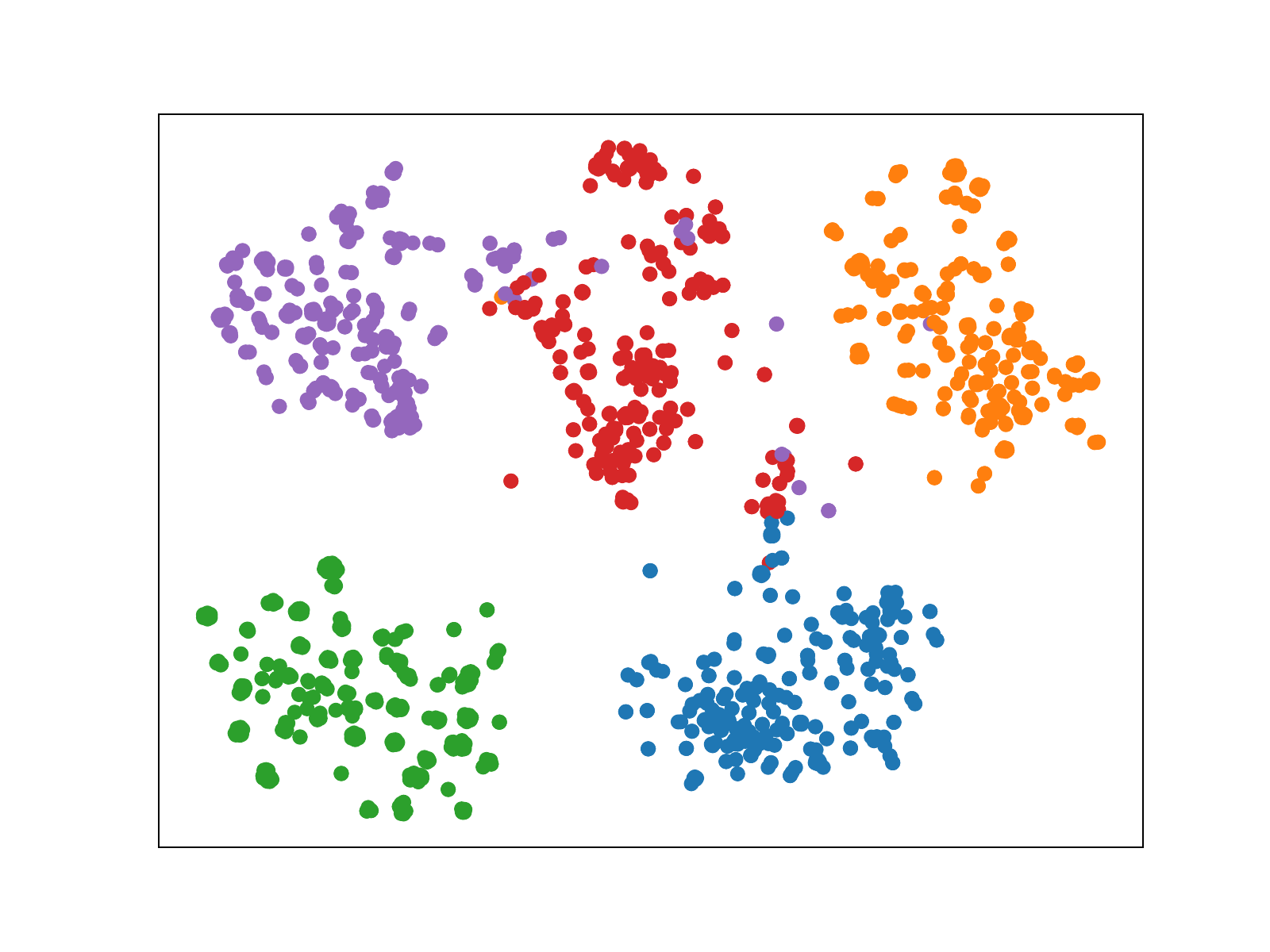}
        \caption{With prototype modulation}
    \end{subfigure}
\vspace{-3mm}
\caption{
\label{fig:tsne_HMDB51}
T-SNE distribution visualization of five action classes on the test set of HMDB51. 
%
The five classes are \textcolor[RGB]{23,118,177}{``dive"}, \textcolor[RGB]{255,122,4}{``climb"}, \textcolor[RGB]{44,160,44}{``climb stairs"}, \textcolor[RGB]{220,45,41}{``drink"} and \textcolor[RGB]{129,77,177}{``dribble"}.
}
\end{figure}

\subsection{Visualization analysis}
%
To further qualitatively analyze our CLIP-FSAR, we visualize the changes of feature distribution without and with prototype modulation in the test phase, as shown in Figure~\ref{fig:tsne}.
By comparing the results, we can observe a clear improvement: the intra-class distributions are more compact, and the inter-classes features are more discriminative after the prototype modulation.
The above findings verify the rationality and superiority of our CLIP-FSAR.

We further visualize the results of the category distribution on the other four datasets, including Kinetics (Figure~\ref{fig:tsne_Kinetics}), UCF101 (Figure~\ref{fig:tsne_UCF101}), SSv2-Full (Figure~\ref{fig:tsne_SSv2-full}), and HMDB51 (Figure~\ref{fig:tsne_HMDB51}).
Five classes are randomly sampled from the test set of these datasets for visualization. 
The results show that after the introduction of prototype modulation, the classes become more distinguishable from each other, and the distribution is tighter within the class.
For instance, as illustrated in Figure~\ref{fig:tsne_Kinetics}, when prototype modulation is not incorporated, the three classes \textcolor[RGB]{23,118,177}{``playing monopoly"}, \textcolor[RGB]{255,122,4}{``hula hooping"}, and \textcolor[RGB]{220,45,41}{``shearing sheep"} are entangled. However, after the addition of prototype modulation, the classes become clearly distinguishable from each other.
In addition, as shown in Figure~\ref{fig:tsne_SSv2-full}, 
the discriminability of the two fine-grained action categories, \textcolor[RGB]{255,122,4}{``Taking [something] out of [something]"} and \textcolor[RGB]{220,45,41}{``Pouring [something] out of [something]"}, is also significantly improved with the incorporation of prototype modulation, thanks to the introduction of powerful textual semantic priors from CLIP to complement visual recognition.

\subsection{Limitations}
%
%
In our CLIP-FSAR, we employ the widely used prompt template, i.e., ``\texttt{a photo of [CLS]}'', as the default setting. 
In Table~\ref{tab:ablation_text_prompt}, we explore the impact of different text prompts and find that different prompt templates perform inconsistently on different datasets, e.g., ``\texttt{[CLS]}'' performs best on  SSv2-Small while worst on the Kinetics dataset. %
It would be valuable work to investigate the design of generic and effective text prompt forms.
In addition,
we mainly focus on the visual aspect to modulate prototypes and do not consider some potential improvements on the text side, such as refining the text features using the visual ones. 
We leave the above discussion for future work.


%
\begin{table}[t]
\vspace{0pt}
\tablestyle{5pt}{1.2}
\centering
\scriptsize
\caption{
Comparison experiments of different prompt templates.
%
}
\vspace{-3mm}
\setlength{
    \tabcolsep}{
    0.5mm}{
    \begin{tabular}
{l|cc|cc}
			
  \multirow{2}{*}{Prompt template}  
& \multicolumn{2}{c|}{{SSv2-Small}}  & \multicolumn{2}{c}{{Kinetics}}  \\
 \multicolumn{1}{c|}{}  
& \multicolumn{1}{l}{1-shot}  & 5-shot 
&  \multicolumn{1}{l}{1-shot}   & 5-shot  \\ \shline

``\texttt{[CLS]}''  & \textbf{53.2}  & \textbf{55.9}   & 87.3   & 91.6    \\
``\texttt{a photo of [CLS]}''  & 52.0  & 55.8   & 87.6   & 91.9    \\
\scriptsize
{``\texttt{a photo of [CLS], a type of action}''}  & 51.6  & 54.6   & \textbf{88.4}   & \textbf{92.1}    \\

%
%
\end{tabular}
}

    \label{tab:ablation_text_prompt}

\end{table}

\section{Conclusion}
\label{sec:conclusion}

In this paper, we propose a CLIP-FSAR method to address the few-shot action recognition problem with the CLIP model, where we fully exploit the multimodal knowledge of CLIP.
%
A video-text contrastive objective is leveraged to adapt the CLIP model to the few-shot video task.
Moreover, we propose to use the text features to adaptively modulate the visual support prototypes by implementing a temporal Transformer.
%
%
Extensive experiments on five commonly used benchmarks demonstrate that our CLIP-FSAR significantly outperforms current state-of-the-art methods. 

\noindent \textbf{Data availability statements.}
The datasets generated during and/or analysed during the current study are available in our open source repository.

\begin{acknowledgements}
This work is supported by the National Natural Science Foundation
of China under grant U22B2053 and Alibaba Group through Alibaba Research Intern Program.

\end{acknowledgements}

{\small
\bibliographystyle{spbasic}
\bibliography{egbib}
}
\end{sloppypar}
\end{document}